\documentclass[opre,nonblindrev]{informs2}

\usepackage{xcolor}
\usepackage{graphicx}
\usepackage{tabularx}
\usepackage[linesnumbered,ruled,vlined]{algorithm2e}

\SetCommentSty{mycommfont}

\SetKwInput{KwInput}{Input}                
\SetKwInput{KwOutput}{Output}              

\usepackage{changepage} 
\usepackage{subfigure}
\usepackage{mathtools}
\usepackage{comment,graphics,epsfig,calc,array, multirow, array,setspace,graphicx,enumitem}
\usepackage{mathrsfs}
\usepackage{amssymb}
\usepackage{pifont}
\newcommand{\ssymbol}[1]{^{\@fnsymbol{#1}}}

\usepackage{natbib}
 \bibpunct[, ]{(}{)}{,}{a}{}{,}%
 \def\bibfont{\footnotesize}%
\TheoremsNumberedThrough     
\ECRepeatTheorems

\EquationsNumberedThrough    

\MANUSCRIPTNO{} 

\newcommand{\teq}{\triangleq}
\newcommand{\1}{\hbox{\rm 1\kern-.35em 1}}
\newcommand{\E}{\mathbb E}
\newcommand{\MS}{\mathscr S}
\newcommand{\MA}{\mathscr A}
\newcommand{\MO}{\mathscr O}
\newcommand{\MP}{\mathscr P}
\newcommand{\MQ}{\mathscr Q}
\newcommand{\MG}{\mathscr G}

\newcommand{\MT}{\mathscr T}
\newcommand{\MI}{\mathscr I}
\newcommand{\MM}{\mathscr M}

\newcommand{\MF}{\mathscr F}

\newcommand{\MV}{\mathscr V}
\newcommand{\MH}{\mathscr H}

\newcommand{\AR}{^\uparrow\kern-.25em \mathbb R^n}
\newcommand{\g}{\mathbf{g}}

\newcommand{\bpi}{\boldsymbol{\pi}}
\newcommand{\bPi}{\boldsymbol{\Pi}}
\newcommand{\bmu}{\boldsymbol{\mu}}
\newcommand{\blambda}{\boldsymbol{\lambda}}
\newcommand{\bH}{\mathbf{H}}
\newcommand{\bC}{\mathbf{C}}

\newcommand{\bh}{\mathbf{h}}
\newcommand{\bS}{\mathbf{S}}
\newcommand{\bs}{\mathbf{s}}
\newcommand{\bpsi}{\boldsymbol{\psi}}
\newcommand{\bPsi}{\boldsymbol{\Psi}}

\newcommand{\bOmega}{\boldsymbol{\Omega}}
\newcommand{\bvartheta}{\boldsymbol{\vartheta}}
\newcommand{\bb}{\mathbf{b}}
\newcommand{\bvarrho}{\boldsymbol{\varrho}}

\newcommand{\un}{\underline}
\newcommand{\ov}{\overline}

\newcommand{\SAV}{\texttt{SAV-Learning}}
\newcommand{\DAV}{\texttt{DAV-Learning}}
\newcommand{\SAVBUC}{\texttt{SAV-Learning-BUC}}
\newcommand{\DAVBUC}{\texttt{DAV-Learning-BUC}}

\begin{document}



\RUNAUTHOR{Soroush Saghafian}
\RUNTITLE{Ambiguous Dynamic Treatment Regimes}

\TITLE{Ambiguous Dynamic Treatment Regimes:\\ A Reinforcement Learning Approach}

\ARTICLEAUTHORS{%
\AUTHOR{Soroush Saghafian}
\AFF{Harvard Kennedy School, Harvard University, Cambridge, MA}
} 

\ABSTRACT{\baselineskip=10pt
A main research goal in various studies is to use an observational data set and provide a new set of counterfactual guidelines that can yield causal improvements. Dynamic Treatment Regimes (DTRs) are widely studied to formalize this process and enable researchers to find guidelines that are both personalized and dynamic. However, available methods in finding optimal DTRs often rely on assumptions that are violated in real-world applications (e.g., medical decision-making or public policy), especially when (a) the existence of unobserved confounders cannot be ignored, and (b) the  unobserved confounders are time-varying (e.g., affected by previous actions). When such assumptions are violated, one often faces ambiguity regarding the underlying causal model that is needed to be assumed to obtain an optimal DTR. This ambiguity is inevitable,  since the dynamics of unobserved confounders and their causal impact on the observed part of the data cannot be understood from the observed data.  Motivated by a case study of finding superior treatment regimes for patients that underwent transplantation in our partner hospital (Mayo Clinic) and faced a medical condition known as New Onset Diabetes After Transplantation (NODAT), we extend DTRs to a new class termed {Ambiguous Dynamic Treatment Regimes  (ADTRs)}, in which the causal impact of treatment regimes is evaluated based on a ``cloud" of  potential causal models.  We then connect ADTRs to Ambiguous Partially Observable Mark Decision Processes (APOMDPs) proposed by \cite{Saghafian2018}, and consider unobserved confounders as latent variables but with ambiguous dynamics and causal effects on observed variables. Using this connection, we  develop two Reinforcement Learning methods termed {Direct Augmented V-Learning} ($\DAV$) and {Safe Augmented V-Learning} ($\SAV$), which enable using the observed data to effectively learn an optimal treatment regime. We establish theoretical  results for  these learning  methods, including (weak) consistency and asymptotic normality. We further evaluate the performance of these learning methods  both in our case study (using  clinical data) and in simulation experiments (using synthetic data). We find promising results for our proposed approaches, showing that they perform well even compared to an imaginary oracle who knows both the true causal model (of the data generating process) and the optimal regime under that model. Finally, we highlight that our approach enables a two-way personalization: obtained treatment regimes can be personalized based on both patients' characteristics and physicians' preferences.\footnote{The author is grateful to Susan Murphy (Harvard), Richard Zeckhauser (Harvard), and Guido Imbens (Stanford)  for their valuable suggestions and comments.}}%


\HISTORY{Version: \today, Forthcoming at {\em Management Science}.}

\maketitle
\baselineskip=18pt
\vspace{-10mm}

\section{Introduction} \label{sec:intro}
In a variety of applications in public policy, governance, medicine, economics, education, energy, and e-commerce,   a main goal is to make better  decisions that are both {\em personalized} and {\em dynamic}. This requires learning from a data set which actions to choose  and when to apply them given the dynamic conditions of each subject (e.g., an individual). One of the main factors that makes this learning process challenging  is that one needs to  estimate the impact of an  alternative sequence of actions  that {\em could have been} used in order to improve outcomes. This requires {\em causal reasoning}, as the estimand---the effect of an alternative sequence of actions---is a {\em counterfactual} quantity \citep[see, e.g.,][]{Murphy2001, Murphy2003, Namkoong2020}.

Dynamic Treatment Regimes (DTRs) have been widely studied for this goal, enabling finding effective alternative policies from  observational data \citep{Robins1986, Robins1997, Murphy2001, Murphy2003, Robins2004, Zhao2015, Zhang2018, Wang2018, Tsiatis2019, Kosorok2019, Luckett2020, Nie2021, Leqi2021}. A DTR is, in essence, a set of rules that prescribe individualized sequence of actions by mapping a subject's history to a series of recommended treatments \citep{Murphy2003, Chakraborty2014, Tsiatis2019, Luckett2020, Xu2020}.

Using available results in finding effective DTRs, however, requires making strong assumptions that might not hold in real-world applications, especially when the data  in hand  is  {\em observational}.  Notably, one needs to assume {\em sequential ignorability}\footnote{This assumption has also appeared in the literature under other names such as ``sequential randomization" \citep{Tsiatis2019} and ``sequential backdoor criterion" \citep{PearlRobins1995}.} \citep{Robins1986, Robins1997, Murphy2001, Murphy2003, Robins2004}, meaning that the data  is rich enough, and hence, unobserved/latent/unmeasured confounding variables either do not exist or their effects can be ignored.   When using observational data sets, this assumption is often violated in many real-world applications.  Even in some secondary analyses of experimental data sets (e.g., those obtained under Micro-Randomized Trials (MRTs)  in some mobile health studies where the goal is to study the effect of users following a treatment regime and not just being assigned to it), various practical challenges (e.g., user habituation,  user engagement, and/or user compliance) may lead to unobserved confounding; see, e.g., \cite{Saghafian2021} for some discussions on scientific challenges in mobile health applications. Furthermore, unobserved confounders are time-varying in most applications: they are themselves affected by the previous actions taken. Adjusting for them, thus, is a perplexing task, making standard approaches for adjustment of confounding erroneous \citep[see, e.g.,][]{Robins2000}.

Correctly adjusting for unobserved time-varying confounding can be managed, if one assumes a specific causal model for the data generating process.\footnote{For example, this can be done under an assumed model for the dynamics of unobserved confounders (e.g., how they are affected by actions taken) and their relationship to observed values (e.g., how unobserved time-varying confounders affect the actions under which data is generated).} Assuming such a model can allow  estimating a distribution for potential trajectories under any alternative decision-making policy (i.e., treatment regime), which is central to estimating its effect.  However, since time-varying confounders are often unobserved, estimating and assuming any such model is subject to significant misspecifications (a.k.a., model ambiguity).   We address this challenge by extending the analyses of DTRs to a new class termed {\em Ambiguous DTRs (ADTRs)}, in which the impact of any sequence of actions is evaluated based on a ``cloud" of potential data generating models as opposed to a single one.  Specifically, we allow for  non-probabilistic ambiguity (a.k.a., Knightian uncertainty) about the true data generating model, while (similar to the literature on DTRs) we assume that under any given potential model, there is a certain probabilistic understanding of how data is generated (see, e.g.,  \cite{Saghafian2018} and Chapter 11  of \cite{Manski2007}  for further discussions,  \cite{Stoy2011} for an axiomatic treatment of statistical decision-making under these conditions, and \cite{Saghafian20161} for an information entropy  view of data-driven decision-making under ambiguity).\footnote{This view of data-driven decision-making under ambiguity has also been shown useful in various applications, including in designing and optimizing queueing systems under model ambiguity \citep{BrenSaghafian2019} and medical decision-making \citep{Boloori2020}.}  This allows for (a) directly taking into account potential model misspecifications when estimating causal impacts, and (b) distinguishing between {\em ambiguity} (lack of knowledge about the true model) and {\em risk} (probabilistic consequences of decisions under a known  model).\footnote{This view is also aligned with that of \cite{Arrow1951}  who stated: ``There are
two types of uncertainty: one as to the hypothesis, which is expressed by saying that the hypothesis is known to belong to a certain class or model, and one as to the future events or observations given the hypothesis, which is expressed by a probability distribution.”.}

In extending DRTs to ADTRs, we are particularly motivated by our various collaborations with our partner hospital, the Mayo Clinic.  In various studies \citep[see, e.g.,][]{Boloori2015, Boloori2020, Munshi2020b, Munshi2020,  Munshi2021}, we have collected data sets from our partner hospital and have examined clinical decisions for patients who undergo a solid organ transplantation  and develop what is known as {\em New Onset Diabetes After Transplantation (NODAT)}.  In practice, physicians often use an intensive amount of an immunosuppressive drug (e.g., tacrolimus) to reduce the risk of organ rejection post-transplant \citep[see, e.g.,][]{Boloori2015, Boloori2020}. Due to a well-established effect known as the diabetogenic effect, this can increase the risk of NODAT, which prompts physicians to use a glucose control drug (e.g., insulin). Learning better ways to prescribe these drugs (e.g., tacrolimus and insulin) in both a personalized and dynamic way  to jointly control risks of  NODAT and organ rejection is not an easy endeavor; the available data sets are only observational, the main health states are hidden \citep[see, e.g.,][]{Boloori2020}, and the existence of unobserved confounders that are time-varying disallow using existing methods.

Our approach in extending DTRs to ADTRs and analyzing them involves the following three main steps. (1) We make use of a utility function that is appropriate under model ambiguity  (instead of  the expected value of outcomes widely used in the literature). (2) We generalize traditional importance sampling methods to accommodate model ambiguity. (3) We connect ADTRs to {\em Ambiguous Partially Observable Mark Decision Processes} (APOMDPs) proposed by \cite{Saghafian2018} by showing that ADTRs can be studied via APOMDPs, which in turn enables us to develop Reinforcement Learning (RL) algorithms capable of  learning optimal treatment regimes from the observed data in effective ways.

The utility function we use is based on a generalization of the traditional {\em maximin expected utility (MEU)} theory (a.k.a.,  Wald's or robust optimization criterion).  The MEU theory assumes that outcomes should  be obtained by maximizing utility with respect to the worst possible member of the ambiguity set (cloud of potential causal models in our setting).  In most applications, using the MEU approach  yields overly conservative decisions \citep[for related discussions, see, e.g.,][and the references therein]{Saghafian2018}, and furthermore, does not allow for representing meaningful human choices such as those of ambiguity seeking individuals established in some behavioral studies \citep[see, e.g.,][]{Bhide2000, Heath1991, Ahn2014}. This was also recognized in the seminal work of \cite{Savage1951} who wrote that this criterion is ``ultrapessimisitic" and ``can lead to absurd conclusion[s]". The generalization we use is known as {\em $\alpha$-maximin expected utility ($\alpha$-MEU)}, which allows for both optimistic and pessimistic views of the world \citep{ArrowHurwitcz, Hurwicz1951a, Hurwicz1951b, Ghiradato2004, Saghafian2018}. Unlike studies that use the MEU criterion, using the $\alpha$-MEU criterion avoids overly conservative decisions by allowing for a controllable {\em pessimism level} (denoted by the parameter $\alpha$) that can take values in [0, 1].

Within the utility theory literature, early studies  \citep[see, e.g.,][]{ArrowHurwitcz, Hurwicz1951a, Hurwicz1951b} provided four axioms that a choice operator must
satisfy. These axioms allowed such studies to show that, under complete ignorance, one can focus merely on two
extreme cases: the best-case and the worst-case. Later studies \citep[see, e.g.,][]{Ghiradato2004, Marinacci2002} further axiomatized preferences under the $\alpha$-MEU criterion  and also highlighted another importance of using the $\alpha$-MEU criterion in decision-making: it allows for differentiating  between the {\em inherent ambiguity} (a property related to the true causal model) and {\em ambiguity attitude} (a property related to the decision-maker). In our study, using the $\alpha$-MEU criterion not only allows us to provide an alternative for the expectation operator---the conventional measure of performance used in the literature surrounding DTRs\footnote{For studies in this literature that consider other measure instead of  the expected value of outcomes, we refer to \cite{Linn2017} and \cite{Wang2018} (quantile performance) and \cite{Leqi2021} (median performance). These studies, however, do not consider model ambiguity, existence of unobserved confounders, or other challenges we aim to address.  While by using the $\alpha$-MEU criterion we primarily generalize the expected value of outcomes, it should be noted that our results can also be used to study generalizations of other measures such as the quantile or median measures.}---but also allows finding treatment regimes that are tailored to the preferences and  attitudes of the decision-maker.

Importantly, this means that our work enables a {\em two-way personalization}: treatment regimes can be personalized based on both the subject's and the decision-maker's characteristics. This is important in various domains such as medicine, where
not only the treatment plan needs to be customized for each patient, but also the physician in charge should be given the ability to include his/her preferences in providing the best course of treatment. Incorporating a physician's preferences is important for many  reasons, including the fact that several behavioral challenges often make it difficult for the physician to follow treatment decisions that are personalized to the patient but not him/her \citep[see, e.g.,][]{Frank2007}.  When using our framework, the physician's preferences in dealing with ambiguous outcomes can be incorporated in various ways. For example, questionnaires similar to those used in preference elicitation methods\footnote{Several studies consider finding treatment regimes that allow shared decision-making between physicians and patients when there are multiple risky (probabilistic)---as opposed to ambiguous (non-probabilistic)---outcomes. Some available methods include set-valued treatment regimes \citep{Laber2014,Lizotte2016}, inverse-preference elicitation \citep{Lizotte2012}, constrained estimation \citep{Linn2015}, and use of item response theory \citep{Butler2018}.} can be designed to first understand the preferences and attitudes of the physician towards ambiguous outcomes, thereby obtaining a small interval (if not a specific value) for  $\alpha$. Using these values of $\alpha$, a small set of corresponding optimal treatment regimes can be presented to the physician for further consideration.  Alternatively, when it is crucial for the physician to follow the treatment regime that provides the maximum robustness and/or efficacy, one can use our framework to find the best treatment regime across all values of $\alpha$ in $[0,1]$. Finally, when other factors beside efficacy or robustness (e.g., cost, availability, or patient consent) need to be considered, one can simply present the set of all treatment regimes that are optimal as $\alpha$ ranges in $[0,1]$, allowing the provider to inspect a broader set of treatment regimes.

 We start our analyses by showing how a generalization of importance sampling methods (a.k.a., inverse-probability-weighting) widely used in the literature \citep[see, e.g.,][]{Robins2000, Precup2000, Murphy2005, Tsiatis2019} can be utilized to find optimal regimes for ADTRs without requiring the dynamics of observed or unobserved variables to be memoryless (i.e., satisfy the Markov property).\footnote{See also \cite{Zhang2019} for more discussions related to fining the optimal treatment regime under model ambiguity without a Markovian structure.} Specifically, we start by generalizing importance sampling methods by allowing sampling across a {\em cloud} of potential data generating models (a.k.a., ambiguity set). We show that under some conditions the resulted method, which we term {\em Generalized Sequential Importance Sampling (GSIS)}, provides a baseline for estimating the causal impact of any dynamic treatment regime, and hence, finding the optimal one.

 When the dynamics of  variables satisfy the Markov property, we connect ADTRs to APOMDPs recently introduced by \cite{Saghafian2018}. APOMDPs generalize traditional POMDPs by allowing model ambiguity.   APOMDPs, however, were proposed without any causal inference application in mind.  In this paper, for the first time, we make use of them through a causal inference lens. Notably, by connecting ADTRs to APOMDPs, we consider time-varying unobserved confounders as dynamic latent states and form dynamic belief distributions over them while allowing ambiguity regarding the true (data generating) causal model.\footnote{Since APOMDPs generalize POMPDs,  our results can also be viewed as generalizations of those in the literature that use a POMDP setting to perform off-policy evaluation \citep[see, e.g.,][and the references therein]{Tennenholtz2019, Xu2020, Bennett2021proximal, Hu2021, Thomas2016}.} We then  make use of known structural results for APODMPs (e.g., piecewise linearity and continuity of the value function) established in the literature \citep{Saghafian2018}, and develop two RL approaches that can provide effective treatment regimes. In developing these RL approaches, as is common, we view the problem of finding an effective treatment regime as an off-policy RL problem. However, in contrast to main RL methods such as Q-Learning (an approximate dynamic programming approach that uses regression to learn the ``quality" function)  and A-Learning (which  tries to learn the ``advantage" function) our approaches try to learn the value function directly.  Thus, roughly speaking they are within the V-Learning methods \cite[see, e.g.,][]{Luckett2020, Xu2020}.  We term our proposed learning algorithms {\em Direct Augmented V-Learning} ($\DAV$) and {\em Safe Augmented V-Learning} ($\SAV$) as they augment the V-Learning methods by (a) making use of the structural properties of the value function, and (b) incorporating model ambiguity (in a direct and safe way, respectively).\footnote{The fact that the structural properties of the value function in APOMDPs is known \citep[see][]{Saghafian2018} is  a main reason we make use of V-Learning as opposed to other RL methods (e.g., Q-Learning or A-Learning). Furthermore, as we will see, a data transformation approach allows using a weight-adjusted version of the Bellman equation, and thereby directly estimating the value function from observed data.}

For our proposed learning approaches, we establish important theoretical results, including weak consistency and asymptotic normality of both the estimated optimal treatment regime and the associate overall gain. To establish these results, we require specific but relatively common ``regularity" conditions,  including conditions on (a)  basic ``complexity" properties of the class of allowable policies (measured by entropy-based versions of the Donsker theorems with bracketing integrals), and (b) absolute regularity of the underlying empirical processes.

We also examine the performance of our proposed approaches by applying them to a clinical data set of over 63,000 observations made of patients who underwent kidney transplantation in our partner hospital and faced NODAT.  We find promising results, indicating that using $\DAV$ and $\SAV$ yields notable improvements over the treatment regime used in practice;  depending on the decision-maker's pessimism level, these improvements are in the ranges (10\%, 42\%) and (10\%, 32\%) for $\DAV$ and $\SAV$, respectively.  Furthermore, we observe that the performance of the $\SAV$ regime is much more robust to the value of the pessimism level (parameter $\alpha$) than that of $\DAV$, and hence, a decisions-maker who uses $\SAV$ does not need to be
worried about the value of $\alpha$ s/he uses in obtaining an optimal treatment regime. We further investigate the performance of our proposed approaches using simulations experiments (synthetic data). Our results show that $\DAV$ and $\SAV$ can improve the observed regime  by an amount that ranges in (1\%,\,37\%) and (1\%,\,8\%), respectively. Furthermore, we make use of our simulation experiments to quantify the robustness of our approaches to model ambiguity, and find that $\DAV$ and the $\SAV$  are able to strongly shield against
model ambiguity: the gain loss under these approaches
compared to an imaginary oracle who knows both the true data generating model and the optimal treatment regime under that model is very low (below 0.6\%), regardless of the value of $\alpha$. Thus, a decision-maker who is facing model ambiguity can make use of our proposed
approaches and obtain a treatment policy that has a similar performance to that of an imaginary decision-maker who knows both the true data generating model and the optimal policy under that model. Finally, our results show that the gain loss compared to such an imaginary decision-maker has a U-shape curve in  the pessimism level: the minimum loss for both
$\DAV$ and  $\SAV$  are obtained at a mid-value of  $\alpha$. This implies that (a) using extreme cases of  $\alpha =0$ (a maximax view) or  $\alpha=1$ (a maximin
view) is almost never {\em robustness-maximizing}, and (b) by viewing $\alpha$ as a tuning parameter (when needed) in our proposed approaches, one can  obtain a treatment regime that performs best across all possible pessimism levels.

In closing this section, we note that our work in incorporating model ambiguity a priori in the analyses not only provides robustness to potential misspecifications, but more broadly, can bridge the gap between two philosophical views of decision-making using causal inference: {\em model-based} and {\em model-free}. The former postulates that any sensible causal reasoning for decision-making  needs to be based on a specific model and set of assumptions in addition to data, while the latter advocates that it needs to rely only on  data. We hope that our work in taking a middle ground and considering a  cloud of models can serve as a step for future research in trying to further bridge the gap between the two. The importance of doing so has its roots in seminal work in Statistical Decision Theory \citep[see, e.g.,][]{Wald1939, Wald1945, Wald1950}, but has also been highlighted in various more recent studies. For example, \cite{Manski2021} emphasizes that ``models can at most approximate actualities"   and highlights that statistical inference for decision-making needs to be performed across all feasible models. Similarly, referring to the famous quote from \cite{Box1979}, \cite{Watson2016}  state that ``statisticians are taught from an early stage that essentially all models are wrong, but some are useful," and stress
that decision-making needs to rely on a set of models that are misspecified (hence ``wrong") but useful in that they can be ``helpful for aiding
actions (taking decisions)."

\section{The Framework} \label{s-framework}

 Throughout the paper, the notation $``\teq"$ is used to differentiate between definitions and equations. For a set $\MT\teq\{1, 2, 3,\cdots, T\}$, the notations $(X_t)_{t\in\MT}$  and $\MT_{\leq t}$ are used to represent the vector $(X_1, X_2,\cdots,X_T)$ and the set $\MT\setminus \{t+1,t+2,\cdots, T\}$, respectively.   All vectors are consider to be in the column format (e.g.,  $(X_t)_{t\in\MT}$ is $|\MT| \times 1$). For any finite set $\Xi\subset \mathbb R$, we let $\Delta_\Xi$  denote the probability simplex induced by $\Xi$.  The notations $\overset{p}{\to}$ and $\overset{d}{\to}$ denote convergence in probability and distribution, respectively. The set $\MI$ represents the interval $[0,1]$.

 We let the observed data be a collection of $n\in\mathbb N$ i.i.d. realizations (called trajectories) of the vector of variables $(O_t, A_t)_{t\in \MT}$. For a realized trajectory, $(o_t, a_t)_{t\in \MT}$, $o_t\in\MO$ is the observation made about a subject (e.g., a patient's observed covariates or an observed health state serving as a summary of them) at time $t\in\MT$,  and $A_t\in\MA$ denotes the action/treatment assigned at time $t\in\MT$, where $\MT$ is the set of time periods (e.g., patients' visits/follow-ups).\footnote{We do not assume that time points are evenly distributed or homogenous across patient trajectories. Importantly, in some applications, the treatment times are random. For simplicity, we assume treatment times are fixed. However, extending our results to scenarios with random treatment times is  relatively straightforward.} For example, in our study of NODAT patients, observations made about each patient ($O_t$) include various test results, demographic information, and other observed risk factors such as diabetes history, body mass index, blood pressure, triglyceride, uric acid, and lipoprotein information (see Table~\ref{Table:Observations}). Actions taken ($A_t$) include low dose (non-aggressive) or high-dose (aggressive) tacrolimus prescriptions as well as information on whether insulin has been used (see Table \ref{Table:Actions}), Finally,  $\MT\teq\{1, 2, 3,\cdots, 12\}$, since patient follow-ups are monthly for a year  after transplantation.

Besides the observed data, there are often unobserved variables that might have affected what is observed in the data. Let $S_t$ denote a summary of them at time $t$, and let $\MS$ be the support of $S_t$. For example, in mHealth applications, $S_t$ might include information relating to the patient's habituation level \citep[see, e.g.,][]{Saghafian2021}  and/or patient true health state, both of which are often unobserved. In our case study of NODAT patients,  $S_t$ is a nine-level variable that summarizes the  unobserved  health state of the patient in terms of both transplantation and diabetes conditions   (see Table \ref{Table:States}). We denote the observable history up to  each time $t\in\MT$ by $\bH^o_t\teq(O_1, A_1, O_2, A_2,\cdots, O_t)$ and let $\MH_t^o$ be the support of $\bH^o_t$. Similarly, we denote the (partially) unobservable history up to  each time $t\in\MT$ by $\bH^u_t\teq(S_1, O_1, A_1, S_2, O_2, A_2, \cdots, S_t, O_t)$ and let $\MH_t^u$ be the support of $\bH^u_t$. It is important to note that in general both variables $S_t$ and $O_t$ depend on the previous treatments. However,  for notational simplicity, we suppress the dependency of $S_t$ and $O_t$ on the vector $(a_t)_{t\in\MT_{\leq t-1}}\teq (a_1, a_2, \cdots, a_{t-1})$.

We assume the latent state summaries $(S_t)_{t\in\MT}$ are such that the immediate gain in each decision epoch depends on the history only through them. This can always be achieved with appropriate definition of  variables $(S_t)_{t\in\MT}$ \citep[see, e.g.,][]{Xu2020}.   For example, in our case study, the immediate gains are based on predefined {\em Quality of Life (QoL)} scores that depend only on  patient summaries defined by  $S_t$ (see Table \ref{Table:Gain}). Thus, we denote the immediate gain at time $t$ through $G_t\teq g (S_t, A_t)\in\mathbb R$, where $g$ is a known function.\footnote{It should be noted that $S_t$, in general,  depends on the history up to time $t$. Thus, $G_t\teq g (S_t, A_t)$ also depends on the history. But this dependence is only through $S_t$, which as noted earlier, can always be achieved with appropriate definition of  summary variables $(S_t)_{t\in\MT}$ \citep[see, e.g.,][]{Xu2020}.}  The set of all possible immediate gains can be denoted by $\MG\teq\{\g^a\in \mathbb R^{|\MS|}:\, a\in \MA\}$, where $\g^a\teq (g(s, a))_{s\in\MS}$.

A treatment regime (hereafter also ``policy" for simplicity) $\blambda\teq(\lambda_t)_{t\in\MT}$ in this setting is a vector of time-dependent mappings from the available history at each time $t$ to the probability simplex induced by actions, $\Delta_\MA$. It defines the probability of assigning each action/treatment at each decision epoch given the available history up to that point. Policies are compared using the overall gain they generate. The overall gain of a policy  $\blambda$  is defined by the discounted sum of immediate gains it generates, which we denote by
\begin{equation}\label{disgain}
\Gamma_T(\blambda)\teq\sum_{t\in\MT} \beta^{t-1} G_t^{\blambda},
\end{equation}
where $\beta\in\MI\setminus\{1\}$ is a discount factor. Similarly, the long-run impact of   $\blambda$ can be analyzed using $\Gamma_\infty(\blambda)\teq\lim_{T\to\infty} \Gamma_T(\blambda)$.\footnote{While we focus on discounted sum of immediate gains, we note that many of our results readily extend to the average overall gains $\bar\Gamma(\blambda)\teq\frac{1}{T}\sum_{t\in\MT} G_t^{\blambda}$, and in particular, to its long-run counterpart $\lim\inf_{T\to\infty} \bar\Gamma(\blambda)$. This is because under some mild conditions $\lim_{T\to\infty} \bar\Gamma(\blambda)= \lim_{T\to\infty} \lim_{\beta\to1} \frac{\Gamma_T(\blambda)}{1-\beta}$.} Here, we shall note that $G_t^{\blambda}$, and hence $\Gamma_T(\blambda)$, should be viewed with a {potential outcomes} lens \citep[for more discussions, see, e.g.,][]{Robins1986, Rubin1986, Angrist1996, Robins1997,Murphy2001}; equivalently, in the language of {do calculus}, $G_t^{\blambda}$ and $\Gamma_T(\blambda)$ should be viewed as $G_t| do(\blambda)$ and $\Gamma_T| do(\blambda)$, respectively \citep[see, e.g.,][]{Pearl2009}. In addition to this, which is implicit in our notation, our notation also implicitly implies {\em consistency}\footnote{This assumption links the counterfactual data with the factual one \citep{Robins1997}, and can be violated if treatment of a subject impacts another subject's variables (e.g., vaccinating a group of individuals may decrease exposure of others to a disease).}, which is a standard assumption in the causal inference literature with  time-varying variables \citep[see, e.g.,][]{Robins1997,Murphy2001} and holds in our motivating study of NODAT patients.  In settings we consider, however, the distribution of $\Gamma_T(\blambda)$ cannot be solely identified from the observed data alone. In fact, there are often a variety of plausible data generating models all agreeing with the observed part of the data, but with different implications about the distribution of $\Gamma_T(\blambda)$.  We let $\MM$ denote the set of all such models (a.k.a., an ambiguity set).\footnote{We defer discussions on how the space of models can be constructed to the numerical experiments section (see, e.g., the discussion under ``Other Details" in the case study). As described there, there are various ways of constructing the set $\MM$, and our analysis does not rely on any specific method or assumption in this regard.} In this view, each given model $m\in\MM$ can be viewed as a rule that, given $\blambda$, imposes a specific probability distribution over the full history $\bH_T^u$. Thus, each given model $m\in\MM$ implies a distribution for $\Gamma_T(\blambda)$, which we denote by $f_m\in\MF$, where $\MF\teq\{f_m: m\in\MM\}$.

 Finally, since the distribution of $\Gamma_T(\blambda)$ varies across the models in $\MM$,  we define a utility function that allows us to compare the performance of different policies. To this end, we make use of  $\alpha$-MEU, which is suitable for decision-making under ambiguity \citep[see, e.g.,][]{Ghiradato2004, Marinacci2002,Saghafian2018}. Specifically, by considering $Y(\blambda)\teq\Gamma_T(\blambda)$ or $Y(\blambda)\teq\Gamma_\infty(\blambda)$ as our main outcome variable of interest, we make use of
\begin{equation}\label{MEU}
MEU_\alpha [Y (\blambda)] \teq \alpha\,\inf_{f^m\in\MF} \E^{f^m} [Y (\blambda)]+ (1-\alpha) \sup_{f^m\in\MF} \E^{f^m} [Y (\blambda)]\ \ \ \ \ \alpha\in\MI,
\end{equation}
as the utility of $Y(\blambda)$, where  $\alpha$ represents the pessimism level and $\E^{f^m}$ denotes the expectation operator with respect to the distribution $f^m$. For example, at $\alpha=1$, (100\% pessimism level),   policies are compared with respect to their worst-case performance. At  $\alpha=0$ (0\% pessimism level), on the other hand, policies are compared with respect to their best case performance. Of note, when $|\MM|=1$, $MEU_\alpha [Y (\blambda)]$ returns the expected value of $Y (\blambda)$, and hence, the utility function in (\ref{MEU}) provides a generalization for the traditional expectation operator that is widely used in the causal inference literature.

We say that the effect of treatment policy $\blambda$ is ``$\alpha$-MEU identifiable," if $\big|MEU_\alpha [Y (\blambda)]\big|<\infty$ and  $MEU_\alpha [Y (\blambda)]$ can be identified given $\MM$. Since a main goal is to learn the  optimal policy, we next define the following notion of optimality in ADTRs, which is a generalization of the traditional notion of optimality used in analyzing DTRs.
\begin{definition} [\textbf{Optimality}]\label{def:opt}
Let $\Lambda$ be the set of all $\alpha$-MEU identifiable policies. We say that a policy ${\blambda}^*\in\Lambda$ is optimal, if with $Y(\blambda)\teq\Gamma_T(\blambda)$, we have
\begin{equation}\label{opttreatment}
MEU_\alpha [Y (\blambda^*)]\geq MEU_\alpha [Y (\blambda)]\hspace{10mm} \forall\blambda\in\Lambda.
\end{equation}
\end{definition}

\begin{remark}[\textbf{Fairness}]\label{remark:fair}
When performance is evaluated under the  traditional expectation operator, it is known that (under some assumptions) the optimal policy assigns treatment (when $|\MA|$=2) only to those subjects who benefit from it: optimizing mean treatment  and conditional mean treatment are equivalent.  However, when using some other measures such as median, this no longer holds: the treatment decision for a given group might depend on outcomes from a different group, creating a ``across-group fairness" concern \citep[see, e.g., ][]{Leqi2021}. This concern is relatively mitigated when using the notion of optimality defined above. This is clear when $|\MM|=1$, since $MEU_\alpha [Y (\blambda)]$ returns the expected value of $Y (\blambda)$. More broadly, it can be seen that, under some conditions, the optimal policy ${\blambda}^*\in\Lambda$ defined in Definition \ref{def:opt} also optimizes $MEU_\alpha [Y (\blambda)]$ after conditioning on subject specific observed variables. For example, if we let $\un m (\blambda^*)\teq \inf_{f^m\in\MF} \E^{f^m} [Y (\blambda^*)]$ and  $\ov m (\blambda^*)\teq \sup_{f^m\in\MF} \E^{f^m} [Y (\blambda^*)]$, assume that  both of these models are in $\MM$, impose similar conditions to those needed when evaluating the mean treatment effect, and evaluate the conditional  $MEU_\alpha$ value after fixing these models (as the worse-case and best-case model, respectively), we can see that $\blambda^*$ satisfies a version of  ``across-group fairness."  Specifically, $\blambda^*$ treats only those subjects who benefit from it in terms of the conditional  $MEU_\alpha$ value, regardless of the outcomes of the other subjects. In addition, when needed, one can further restrict the set of allowable policies $\Lambda$ to those that satisfy some desirable fairness attributes.  However, studying fairness and what should be considered as ``fair" is outside the scope of this work, especially since measuring fairness under {\em ambiguity} is more complex than that under {\em risk};  see the discussion in the Introduction that highlights the difference between the two as well as more in-depth discussions in prior work \citep[see, e.g.,][] {Saghafian20161, Saghafian2018}.   In what follows, we simply focus on data-driven ways of findings a policy ${\blambda}^*\in\Lambda$ that is optimal based on Definition \ref{def:opt}.
\end{remark}

To perform our analyses,  it is useful to differentiate between the policy under which the data have been generated (hereafter, the ``behavior policy") and the policy that we would like to evaluate and recommend (hereafter, the ``evaluation policy"). The behavior policy denoted by $\blambda^b\teq(\lambda_t^b)_{t\in\MT}$  is  a vector of time-dependent mappings $\lambda_t^b: \MH^u_t\to\Delta_\MA$ whereas the evaluation policy denoted by $\blambda^e\teq(\lambda_t^e)_{t\in\MT}$  is  a vector of time-dependent mappings $\lambda_t^e: \MH^o_t\to\Delta_\MA$. An important difference between the evaluation and the behavior policies relates to a condition known as {\em sequential ignitability}\footnote{See also the {\em sequential backdoor} criterion \citep{PearlRobins1995}.} \citep[see, e.g.,][]{Robins1986, Robins1997, Murphy2001, Murphy2003, Robins2004}, which we define next.

\begin{definition}[\textbf{Sequential Ignorability}] For any policy $\blambda\teq(\lambda_t)_{t\in\MT}$, let $\bH^{o,m}_t (\blambda)\teq(O^{m, \blambda}_1, A^{m, \blambda}_1, O^{m, \blambda}_2, A^{m, \blambda}_2,\cdots, O^{m, \blambda}_t)$ denote the observable history up to time  $t\in\MT$, generated under  $\blambda$ and model $m\in\MM$.  We say that $\blambda$ satisfies sequential ignorability under model $m\in\MM$, if for all $t\in\MT$, the action generated
by $\lambda_t$ is independent of the collection of potential outcomes $(G^{m, \blambda'}_{t}, O^{m, \blambda'}_{t+1}, G^{m, \blambda'}_{t+1}, O^{m, \blambda'}_{t+2}, \cdots, G^{m, \blambda'}_{T})_{\blambda'\in\Lambda}$ conditional on $\bH^{o,m}_t (\blambda)$.
\end{definition}

In essence, this definition requires, for each time $t\in\MT$ and given $\bH^{o,m}_t (\blambda)$,   the action generated by $\lambda_t$ to be independent of $(G^{m}_{t}, O^{m}_{t+1}, G^{m}_{t+1}, O^{m}_{t+2}, \cdots, G^{m}_{T})$ values that can be obtained when following any feasible sequence of actions $(a_t)_{t\in\MT}$. Both by this definition and naturally,  any evaluation policy  $\blambda^e\teq(\lambda_t^e)_{t\in\MT}$ (where $\lambda_t^e: \MH^o_t\to\Delta_\MA$) satisfies sequential ignorability under any model $m\in\MM$, because it is only a function of the observed history and possible exogenous randomness (it maps $\MH^o_t$ to $\Delta_\MA$by definition).  In contrast, a behavior policy $\blambda^b\teq(\lambda_t^b)_{t\in\MT}$ (where $\lambda_t^b: \MH^u\to\Delta_\MA$)  may or may not satisfy this condition, since it might depend on unobservable confounders (variables in $(S_t)_{t\in\MT}$ that affect both the gain and the actions selected by $\blambda^b$). In fully randomized experiments (e.g., Micro Randomized Trials), the behavior policy  may satisfy sequential ignorability. However, when the data are observational, it is often impossible to test whether the behavior policy satisfies this assumption, and in addition, it is  highly likely that this assumption does not hold. Finally, we note that while the behavior policy is often known in randomized experiments, it might need to be estimated when using observational data (unless specific known treatment protocols or standard are fully followed).

\subsection{Analyzing ADTRs via Generalized Sequential Importance Sampling (GSIS)}
We now show that, under some conditions, an optimal policy for an ADRT can be found using  a generalized version of sequential importance sampling, which we term  {\em Generalized Sequential Importance Sampling (GSIS)}. While allowing for {\em ambiguity} (not risk, which is the focus of the existing literature), GSIS assigns weights under each model and sequentially adjusts the trajectory probabilities that occur under a given evaluation policy compared to those observed in the data set. Of note, we use GSIS in this section to study ADTRs that do not satisfy any Markovian (a.k.a., memoryless) property regarding the dynamics of the underlying variables. In the next section, we show how the analyses of ADTRs can be simplified when such dynamics satisfy a Markovian structure. Notably, the results presented in this section provide a building block for the algorithms and the theoretical findings established in the next sections.

To present GSIS, we first suppress the dependencies to the underlying model by assuming the model is fixed. Consider an evaluation policy $\blambda^e$,  and let $\bH^o_t (\blambda^e)$ be the history that will be observed under $\blambda^e$ up to time $t$. Also, denote by $\lambda_t^e(A_t|\bH^o_t (\blambda^e))$ the probability that actions $A_t$ is chosen under $\blambda^e$ when the observed history is $\bH^o_t (\blambda^e)$. Furthermore,  while the behavior policy is not known (e.g., due to its potential dependency on unobserved variables), we can observe the {\em marginalized} probabilities of action selection under the behavior policy, which we denote by $\lambda_t^b(A_t|\bH^o_t (\blambda^b))$. These allow us to define importance sampling weights
\begin{equation}\label{eq:W}
w_t (\blambda^e)\teq \frac{\lambda_t^e(A_t|\bH^o_t (\blambda^e))}{\lambda_t^b(A_t|\bH^o_t (\blambda^b))} \ \ \ \ \forall t\in\MT.
\end{equation}

Proposition \ref{prop:SIS}  establishes that, under some conditions, the optimal policy for an ADTR governed by a set of models $\MM$ can be found via GSIS. Specifically,  an $MEU_\alpha$--unbiased estimator of  $\Gamma_T(\blambda^e)$ is obtained in Proposition \ref{prop:SIS}, where the notion of $MEU_\alpha$--unbiased estimation is defined below.

\begin{definition}[\textbf{$MEU_\alpha$--Unbiasedness}]\label{def:unbias}
An estimator $\hat Y$ of an outcome variable of interest $Y$  is said to be $MEU_\alpha$--unbiased if, and only if, $MEU_\alpha[\hat Y]=MEU_\alpha[Y]$ for any $\alpha\in\MI$.
\end{definition}

To establish an $MEU_\alpha$--unbiased estimator of  $\Gamma_T(\blambda^e)$,  we also need to make sure that the evaluation and the behavior policies sufficiently {\em overlap}. Specifically, we need to ensure that these policies overlap almost surely (defined below).

\begin{definition}[\textbf{Almost Sure Overlap}] We say that the evaluation and the behavior policy almost surely overlap, if ${\lambda_t^b(a_t|\bH^o_t (\blambda^b))}>0$ whenever ${\lambda_t^e(a_t|\bH^o_t (\blambda^e))}>0$ a.s. over $\bH^o_t(\blambda^b)$ and $\bH^o_t(\blambda^e)$ for all $t\in\MT$ and $a_t\in\MA$.
\end{definition}

Intuitively, the evaluation and the behavior policy need to overlap to ensure that trajectories obtained under the behavior policy are to some extent informative about the trajectories under the evaluation policy. When the evaluation and the behavior policy almost surely overlap, the importance sampling weights defined in \eqref{eq:W} are well-defined for all $t\in\MT$ (except perhaps on histories that might happen with probability zero).

\begin{proposition}[Generalized Sequential Importance Sampling (GSIS)]\label{prop:SIS} Suppose that the evaluation and behavior policies (a)~satisfy sequential ignorability under all models $m\in\MM$, and (b)~almost surely overlap. Then,  for any  $\alpha\in\MI$, we have
\begin{equation}\label{prop:SISADTR}
MEU_\alpha\big[\Gamma_T(\blambda^e)\big]=MEU_\alpha\big[\Gamma_T(\blambda^b) \prod_{t\in\MT} w_t(\blambda^e)\big],
\end{equation}
and hence, $\hat \Gamma_T(\blambda^e)\teq \Gamma_T(\blambda^b) \prod_{t\in\MT} w_t(\blambda^e)$ is an $MEU_\alpha$--unbiased estimator of  $\Gamma_T(\blambda^e)$.
\end{proposition}

The proof of Proposition \ref{prop:SIS} is developed by understanding how the impact of an evaluation policy can  be first analyzed for any given (a) sequence of actions, and (b) model $m\in\MM$ under which the data might be generated (see Lemma \ref{lemma:actionseq} in Appendix A).

Of note, Proposition \ref{prop:SIS} also provides a partial way of characterizing the optimal evaluation policy, since it provides a way of estimating $MEU_\alpha\big[\Gamma_T(\blambda^e)\big]$ under any given evaluation policy.  That is, using this proposition and optimizing $MEU_\alpha\big[\Gamma_T(\blambda^e)\big]$ over a given set of policies (which for use in practice might be restricted to those satisfying desirable attributes such as fairness or interpretability) can shed light on the optimal evaluation policy. However, Proposition \ref{prop:SIS} provides only a partial way of characterizing the optimal evaluation policy, because  analyzing ADTRs often requires considering a behavior policy that might not satisfy sequential ignorability (at least under some models in $\MM$).  Therefore, we next study scenarios in which the behavioral policy does not fully satisfy sequential ignorability, but satisfies it {\em to some extent}. This entails limiting the impact of unobserved confounders (which make the probability of observing certain trajectories in the observed data biased compared to what would have happen if we could observe unobservables)  on the behavior policy under each model. In limiting the impact of unobserved confounders on the behavior policy, we are mainly motivated by extending the analyses of confounding in causal inference \citep[see, e.g.,][]{Rosenbaum2002} from a traditional  setting in which  $|\MT|=1$, the treatment variable is binary $|\MA|=2$, and there is no model ambiguity $|\MM|=1$, to ADTRs in which these restrictions are all relaxed. Two notable challenges in doing so are: (1) since future actions depend on the history, a confounding decision/treatment in any period can make future decisions confounding as well; (2)   since the trajectory probabilities depend on the underlying model, the impact of unobserved confounding depends on the underlying model. We next introduce the notion of {\em bounded unobservable confoundedness}, which we define using the likelihood ratios of treatment propensities (functions $\ell (\cdot)$ in the following definition). This, in turn, allows us to provide a version of GSIS under bounded unobservable confoundedness (Proposition \ref{prop:buc}).

\begin{definition}[\textbf{Bounded Unobservable Confounding (BUC)}]\label{def:boundedconf} We say that the behavioral policy satisfies Bounded Unobserved Confounding (BUC) under  a model $m\in\MM$, if there exist constants $\eta_t^m\in[1,\infty)$ such that
\begin{equation}
(\eta^m_t)^{-1}\leq\frac{\ell (a_t,a'_t, \bH_t^{o,m}, \bS^m_t=\bs)}{\ell (a_t,a'_t, \bH_t^{o,m}, \bS^m_t=\bs')}\leq \eta^m_t
\end{equation}
a.s. over observable history $\bH_t^{o,m}$,   for all $t\in\MT$, $a_t,a'_t\in \MA$, $\bs, \bs'\in\MS^{(T)}$, where $\bS^m_t\teq(S^m_t)_{t\in\MT_{\leq t}}$ and
$$ \ell (a_t,a'_t, \bH_t^{o,m},  \bS^m_t=\bs)\teq \frac{\lambda^b_t (a_t|\bH_t^{o,m}, \bS^m_t=\bs)}{\lambda^b_t (a'_t|\bH_t^{o,m},\bS^m_t=\bs)}.$$
\end{definition}
The above definition bounds the impact of the vector of the unobservable confounder variables, $\bS^m_t$,  in each period. In Lemma \ref{lemma:ec:boundedconf} (Appendix B) we show that this definition results in
\begin{equation}\label{eq:BUCEquivalent}
(\eta^m_t)^{-1}\leq\frac{\lambda^b_t (a_t|\bH_t^{u,m})}{\lambda^b_t (a_t|\bH_t^{o,m})}\leq \eta^m_t\ \ \ \ \ a.s.
\end{equation}
over $\bH_t^{o,m}$ and $\bH_t^{u,m}=(\bH_t^{o,m}, \bS^m_t)$ for all $t\in\MT$ and $a\in\MA$. Thus, benefiting from the observed history (as opposed to the unobserved one) and making use of marginalized  propensities $\lambda^b_t (a_t|\bH_t^{o,m})$
as an estimate of  the true treatment propensities $\lambda^b_t (a_t|\bH_t^{u,m})$ will not be unboundedly misleading. The results provided in the following proposition   are analogous to {\em design sensitivity} analyses \citep[see, e.g.,][]{Rosenbaum2010} in static (i.e., $T=1$) settings, where the idea is to examine how much propensity odds need to vary so that the gained causal understanding becomes  invalid \citep[see, also, ][]{Kallus2020Conf, Kallus2021}. This proposition can also be viewed as a generalization of some of the available bounds in the literature of DRTs \citep[see, e.g., Lemma 2 of ][]{Namkoong2020}, since such bounds can be obtained from our results under the special case of $|\MM|=1$.  Furthermore, we note that while these bounds can be conservative (hence, not useful) for the purpose of estimating the mean performance, they are relatively suitable for estimating the $MEU_\alpha$ value of it (see part (ii) of  Proposition \ref{prop:buc} as well as the learning approaches discussed in Section \ref{sec:extension}).

\begin{proposition}[GSIS under Bounded Unobservable Confounding]\label{prop:buc}
Suppose the behavior policy  satisfies BUC under all models $m\in\MM$. If the evaluation policy  satisfies sequential ignorability under all models $m\in\MM$, and it overlaps with the behavior policy almost surely, then:
\begin{itemize}[leftmargin=1cm,align=left]
\item[(i)] Under each model $m\in\MM$ we have:
$$\E^m\big[\Gamma_T(\blambda^b) \prod_{t\in\MT} \un w^m_t(\blambda^e)\big] \leq \E^m\big[\Gamma_T(\blambda^e)\big]\leq \E^m\big[\Gamma_T(\blambda^b) \prod_{t\in\MT} \ov w^m_t(\blambda^e)\big],$$
where
$$\un w^m_t(\blambda^e)\teq w_t(\blambda^e)\, \Big(({\eta^m_t})^{-1}\, \1_{\{\Gamma_T(\blambda^b) >0\}}+ {\eta^m_t}\, \1_{\{\Gamma_T(\blambda^b) <0\}}\Big),$$
and
$$\ov w^m_t(\blambda^e)\teq w_t(\blambda^e)\, \Big(({\eta^m_t})^{-1}\, \1_{\{\Gamma_T(\blambda^b) <0\}}+ {\eta^m_t}\, \1_{\{\Gamma_T(\blambda^b)>0\}}\Big).$$
\item[(ii)] For any $\alpha\in\MI$, there exists $\tilde\alpha\in\MI$ such that
$MEU_\alpha\big[\Gamma_T(\blambda^e)\big]= f(\tilde\alpha)$,
where $f (\tilde\alpha)\teq \tilde\alpha\, MEU_\alpha\big[\Gamma_T(\blambda^b) \prod_{t\in\MT} \un w^m_t(\blambda^e)\big]+ (1-\tilde\alpha) MEU_\alpha\big[\Gamma_T(\blambda^b) \prod_{t\in\MT} \ov w^m_t(\blambda^e)\big].$
\end{itemize}
\end{proposition}

Similar to Proposition  \ref{prop:SIS}, part (ii) of Proposition  \ref{prop:buc} provides a way of finding the optimal evaluation policy, since it characterizes the causal impact of any such policy.  Whereas Proposition 1 requires the behavior policy to satisfy sequential ignorability---an
unrealistic assumption in most applications---Proposition \ref{prop:buc} only requires the unobserved variables
to have a bounded impact. Importantly, however, Proposition \ref{prop:SIS} directly provides an $\alpha$-MEU unbiased estimator, but  Proposition \ref{prop:buc} does so subject  to a tuning parameter $\tilde\alpha$. Specifically, in part (ii) of Proposition \ref{prop:buc}, the function $f$ can be computed using only observed data. This, in turn, resolves the issue that the outcome of interest under the evaluation policy as well as the time-varying confounders needed to estimate it are unobservable. However, to use Proposition \ref{prop:buc} part (ii), one needs to tune the parameter $\tilde\alpha$. Since $f$ is a decreasing function and $\tilde\alpha\in\MI$, tuning  $\tilde\alpha$ can be done in an structured way.  For example, in practice, one is often  interested in evaluating policies that are known to be better than the behavior policy. Thus, we  have $f (\tilde\alpha)\geq MEU_\alpha\big[\Gamma_T(\blambda^b)\big]$, implying that one can start tuning $\tilde\alpha$ using the  threshold value $\tilde\alpha^*\teq \min\bigg\{f^{(-1)} \Big(MEU_\alpha\big[\Gamma_T(\blambda^b)\big]\Big), 1\bigg\}$ and only consider values of $\tilde\alpha$ that are in $[0,\tilde\alpha^*]$. More importantly, it should be noted that the parameters $(\eta_t^m)_{t\in\MT,m\in\MM}$ are design sensitivity parameters. Specifically, for any $\epsilon>0$, they can be chosen so that $f(0)-f(\tilde\alpha^*)<\epsilon$. Since $f(0)-f(\tilde\alpha^*)\geq 0$, this allows one to use any $\tilde\alpha$ in $[0,\tilde\alpha^*]$ and obtain an {\em approximate} unbiased $MEU_\alpha$ estimator for $\Gamma_T(\blambda^e)$ with a guaranteed approximation error of $\epsilon$. While this provides an approximation method, two limitations are noteworthy: (1) obtaining an exact value for  $\tilde\alpha$ (and hence, an exact unbiased $MEU_\alpha$ estimator for $\Gamma_T(\blambda^e)$) can be challenging, and (2)  the bounds in part (i) of Proposition  \ref{prop:buc} can be conservative, since in general they may diverge exponentially in $T$. Nonetheless, as described above, one can obtain an  $\epsilon$-approximate unbiased $MEU_\alpha$ estimator for $\Gamma_T(\blambda^e)$.

Finally, we note that one can extend Propositions \ref{prop:SIS} and \ref{prop:buc} to provide doubly robust estimators\footnote{For related studies on doubly robust estimators, we refer to \cite{Bang2015, Jiang2015, Thomas2016, Kallus2020, Athey2021}, and the references therein.}  to account for the fact that, under each given model $m\in\MM$, the variance of an importance sampling based estimator can be high. Such an extension is, however, not that useful in our work, because we are (a) directly allowing for  a cloud of models, and (b) using $MEU_\alpha$ of the outcome variable as opposed to its expected value (the criterion used in the studies related to doubly robust estimation). Instead, we next develop two RL methods based on our results, and establish that they have suitable asymptotic behavior, including consistency and asymptotic normality. We also test their performance directly using both a clinical data set and simulation experiments, and find that our proposed learning methods provide strong robustness to model ambiguity (see, e.g., Section~\ref{sec:robsutness}).

\section{Analyzing ADTRs via APOMDPs}\label{sec:APOMDP}

In this section, we show that a tractable way of analyzing ADTRs is via  APOMDPs. Specifically, analyzing ADTRs  via  APOMDPs enables (a) considering unobserved variables as latent time-varying states while allowing for model ambiguity, and (b) developing effective RL methods.\footnote{For other approaches in modeling confounders as hidden states see, e.g., \cite{Bennett2021, Xu2020}, and the references therein.}

\begin{figure}[t]
  \begin{center}
  \includegraphics[scale=0.65]{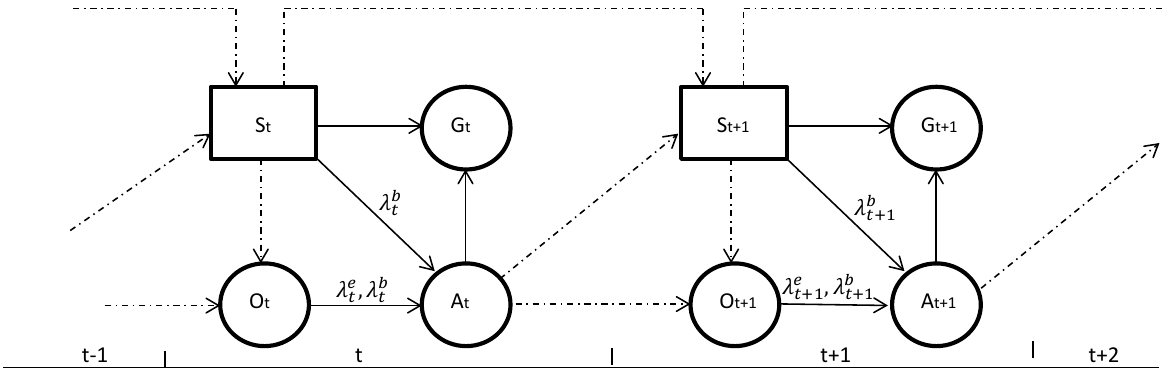}\vspace{5mm}
  \caption{\scriptsize DAG representation of  APOMDPs. {\em Circles:} observable variables; {\em Rectangles:} unobservable variables; {\em Solid arrows:} unambiguous causal mechanisms; {\em Dashed arrows:} ambiguous causal  mechanisms. [Note: using $\bpi_t\teq\nu_t(\bh_t^o)$---a belief distribution over the latent states---allows working with an equivalent DAG representation where $\bpi_t$ becomes the observed state at time $t$.]}\label{fig:apomdpascm}
    \end{center}\vspace{-4mm}
\end{figure}

An APOMDP \citep{Saghafian2018} can be represented via the Directed Acyclic Graph (DAG) depicted in Figure \ref{fig:apomdpascm}.  The ambiguous mechanisms in this figure represent causal relationships that cannot be quantified from the data alone. The main  assumption needed to represent an ADTR via an APOMDP is that the dynamics of the variables is Markovian. In various applications, it is often possible to transform data so that this assumption holds \citep[see, e.g.,][]{Xu2020}. Specifically, while the observed history $\bH^o_t$ grows over time, we can assume  that there are summary functions $\nu_t:\MH_t^o\to\Delta_\MS$ such that $\bpi_t\teq\nu_t(\bh_t^o)$ (a belief distribution over the latent states) is a sufficient statistics.\footnote{For typical POMDPs and APOMDPs, it is known that the belief distribution over latent states can serve as a sufficient statistics \citep[see, e.g.,][and the references therein]{Saghafian2018, Boloori2020,Saghafian20194}. In Remark \ref{remark:BUC} and Section \ref{sec:extension}, we further discuss handling cases where this might not hold. We also refer interested readers to Assumption 1 in \cite{Tennenholtz2019}, which establishes the existence of a sufficient statistics as one sufficient condition for unbiasedness of importance sampling in POMDPs, but highlight that our focus is on APOMDPs as opposed to POMDPs.}

 Using the belief distribution $\bpi_t$, we can work with  transformed policies:  we can consider  $\bmu^e\teq\big(\mu^e_t(\bpi_t)\big)_{t\in\MT}$ and $\bmu^b\teq\big(\mu^b_t(\bpi_t)\big)_{t\in\MT}$ as the evaluation and behavior policies, respectively,  where $\mu^e_t, \mu^b_t: \Delta_\MS\to\Delta_\MA$.  We denote the probability that an action $a_t$ is applied at time $t$ (when the belief distribution is $\bpi_t$) under these transformed evaluation and behavior policies by   $\mu^e_t(a_t|\bpi_t)$ and   $\mu^b_t(a_t|\bpi_t)$, respectively. In what follows, we first define the class of APOMDPs and then develop two RL algorithms that enable finding the optimal policy by effectively learning the causal impact of any given evaluation policy.

As defined in \cite{Saghafian2018},  a time-homogenous APOMDP is an extension of the classical POMDPs, and can be defined by the tuple  ($\alpha$, $\beta$, $\MS$, $\MO$, $\MA$, $\MG$, $\MP$, $\MQ$). The notation used in the first part of this tuple is as introduced earlier.  $\MP$ and $\MQ$ are the sets of possible transition probability matrices   with respect to (latent) states and observations, respectively \citep{Saghafian2018}. These sets define the ambiguous causal mechanisms depicted in Figure \ref{fig:apomdpascm}.

To simplify the analyses, we can index members of the set $\MP\times\MQ$ using $\MM$ so that each $m\in\MM$ represents  a specific (unambiguous) POMDP model. In particular, associated with each  $m\in\MM$  is  a set of the form $P_m \times Q_m$ with $P_m\in\MP$ and $Q_m\in\MQ$ denoting the set of state and observation transition probabilities under model $m$, respectively \citep{Saghafian2018}. In this setting, (a) $P_m\teq\{P_m^{a}: a\in \MA\}$, where for each $a\in \MA$, $P_m^a\teq [p^a_{ij}(m)]_{i,j\in\MS}$ is an $|\MS|\times |\MS|$ matrix with $p^a_{ij}(m)\teq Pr\{j|i,a,m\}$ denoting the probability that the (latent) state moves to $j$ from $i$ under action $a$ and model $m$, and (b)  $Q_m\teq\{Q_m^a: a\in \MA\}$,  where for each $a\in \MA$, $Q_m^a\teq [q^a_{jo}(m)]_{j\in\MS, o\in\MO}$ is an $|\MS|\times |\MO|$ matrix with $q^a_{jo}(m)\teq Pr\{o|j,a,m\}$ denoting the probability of observing $o$  under action $a$ and model $m$ when the (latent) state is $j$ \citep{Saghafian2018}.

If $\MM$ was a singleton with its only member being $m$, the optimal gain and policy for any $t\in\MT$ and $\bpi \in \Delta_\MS$ could be obtained by a traditional POMDP Bellman equation (along with the terminal condition $V_0^m(\bpi)\teq 0$):
\begin{equation}\label{traditional}
V^m_t(\bpi)=\max_{a\in \MA}\Big\{\bpi'\g^a+\beta\sum_{o\in\MO} Pr\{o|\bpi,a, m\}V^m_{t-1}(T(\bpi, a, o, m))\Big\},
\end{equation}
where $V^m_t(\bpi)$ denotes the value function under model $m$ when the belief distribution is $\bpi$ and there are $t$ periods to go,  `` $'$ " represents the transpose operator,   $Pr\{o|\bpi,a, m\}=\sum_i\sum_j \pi_i p_{ij}^a (m) q_{jo}^a (m)$, and the belief updating  operator $T:\, \Delta_\MS\times \MA\times\MO\times\MM\to\Delta_\MS$  is defined by the Bayes' rule (in the matrix form):
\begin{equation}\label{beliefupd}
T(\bpi, a, o, m)=\frac{\big(\bpi'P^a_m Q^a_m(o)\big)'}{Pr\{o|\bpi, a, m\}},
\end{equation}
with $Q^a_m(o)\teq\text{diag} (q^a_{1o}(m), q^a_{2o}(m),\ldots, q^a_{no}(m))$ denoting the diagonal matrix made of the $o$th column of $Q^a_m$ \citep{Saghafian2018}.

Unlike POMDPs, in AMPOMDs $\MM$ is not  a singleton. However, it is shown in \cite{Saghafian2018} that  the APOMDP value function, a model independent function which we denote by $V_t(\bpi)$,  can still be obtained using dynamic programming. Furthermore,   the underlying Bellman operator in the APOMDP is a contraction mapping with modulus $\beta$ on a complete metric space (under some mild conditions), which in turn allows analyzing the APOMDP value function in infinite-horizon settings as the limit of its finite-horizon version. More importantly, \cite{Saghafian2018} establishes some structural properties for the value function of the APOMDP (e.g, piecewise linearity and continuity in $\bpi$). In the next section, we make use of these structural properties to develop effective RL  approaches (termed Augmented V-Learning).  We start our analyses by first developing suitable algorithms for learning the value function in POMDPs (i.e., when $|\MM|=1$), and then show how they can  be extended to learn the APOMDP value function.

\section{Augmented V-Learning for POMDPs and APOMDPs}
\subsection{Augmented V-Learning  for POMDPs}\label{sec:AVPOMDP}

To develop our results, we require that the behavior policy, $\bmu^b$, satisfies {\em positivity} defined below.

\begin{definition}[\textbf{Positivity}]
We say that a policy $\bmu\teq (\mu_t)_{t\in\MT}$ satisfies positivity, if  there exists a constant $c_0>0$ such that $\mu_t (a_t| \bpi_t)\geq c_o$ for all $t\in\MT$,  $\bpi_t\in\Delta_\MS$, and $a_t\in\MA$.
\end{definition}

Positivity implies that  all actions have a positive chance of being selected (appear in the observed data) regardless of the belief.  The behavior policy, $\bmu^b$, automatically satisfies  positivity  when the data are collected based on a randomized trial. When using observational data this assumption is  sensible, because  inference involving treatment patterns (using action $a_t$ when the belief is $\bpi_t$)  that cannot occur in the observational study requires further knowledge and assumptions \citep{Murphy2001}. If the behavior satisfies positivity, we can establish the following result (see also Lemma 4.1 of  \cite{Murphy2001} and  Lemma 2.1 of \cite{Luckett2020} for related results in settings with fully observable states).

\begin{proposition}[Weight-Adjusted Bellman Equation]\label{prop:POMDPest}
Suppose $|\MM|=1$ and denote the only member of $\MM$ by $m$. If $\bmu^b$ satisfies positivity and sequential ignorability, then for any policy $\bmu^e$, the finite-horizon value function satisfies the weight-adjusted Bellman equation

\begin{equation}\label{eq:POMDPest1}
V_{T-t+1}^{m,\bmu^e} (\bpi_t)= \E^m \bigg[\frac{\mu_t^e(A_t|\bPi^m_t)}{\mu_t^b(A_t|\bPi^m_t)}\Big[G_t+ \beta\, V_{T-t}^{m,\bmu^e}  (T(\bPi^m_t, A_t, O_t, m))\Big]\Big| \bPi^m_t=\bpi_t \bigg],
\end{equation}
for all $t\in\MT$ and $\bpi_t\in\Delta_\MS$, where $\bpi_t$ is considered as a realization (of a model dependent random variable denoted by $\bPi^m_t$) and $V_{0}^{m,\bmu^e} (\bpi)\teq 0$. Therefore, for any function $\phi$ defined on $\Delta_\MS$, and for all $t\in\MT$, we have:
\begin{equation} \label{eq:POMDPest2}
\E^m \bigg[ \frac{\mu_t^e(A_t|\bPi^m_t)}{\mu_t^b(A_t|\bPi^m_t)}\Big[G_t+ \beta\, V_{T-t}^{m,\bmu^e}  (T(\bPi^m_t, A_t, O_t, m))- V_{T-t+1}^{m,\bmu^e} (\bPi^m_t)\Big] \phi (\bPi^m_t)\bigg]=0.
\end{equation}
\end{proposition}

The importance of Proposition \ref{prop:POMDPest} (which is built on the importance sampling results of the previous section)  is that it allows us to empirically estimate the value function under any evaluation policy, and hence, learn the optimal policy. Specifically, using the data, we can make use of the sample-average version of \eqref{eq:POMDPest2}:
\begin{equation} \label{eq:POMDPest3}
\E^{\mathbb P} \Bigg[ \sum_{t\in\MT}\bigg[ \frac{\mu_t^e(A_t|\bPi^m_t)}{\mu_t^b(A_t|\bPi^m_t)}\Big[G_t+ \beta\, V_{T-t}^{m,\bmu^e}  (T(\bPi^m_t, A_t, O_t, m))- V_{T-t+1}^{m,\bmu^e}  (\bPi^m_t)\Big] \phi (\bPi^m_t)\bigg]\Bigg]=0,
\end{equation}
where $\E^{\mathbb P}$ denotes average with respect to the empirical probability measure.\footnote{For a random variable $X$ with $n$ observed values denoted by $x_1, x_2,\cdots, x_n$, $\E^{\mathbb P}[X]\teq n^{-1} \sum_{i=1}^n x_i$. Similarly, for a function $f$, $\E^{\mathbb P}[f(X)]\teq n^{-1} \sum_{i=1}^n f(x_i)$.}  It is important to note that while we are using sample-average in \eqref{eq:POMDPest3}, the result still depends on the assumed $m$, because while the sequence $\{(A_t, O_t)\}_{t\in\MT}$ is observable to us, to form the sequence $\{\bPi^m_t\}_{t\in\MT}$, we need to have an assumed model. That is, due to the existence of unobserved variables, the empirical measure alone is {\em insufficient} for our goal.

\begin{remark} [\textbf{Weight Adjustment under BUC}]\label{remark:BUC}
The weight-adjusted results \eqref{eq:POMDPest1}-\eqref{eq:POMDPest3} is obtained assuming that $\bmu^b$ satisfies sequential ignorability. It should be noted that this assumption is milder than assuming that the original behavior policy satisfies sequential ignorability. This is especially the case, because $\bmu^b\teq\big(\mu^b_t(\bpi_t)\big)_{t\in\MT}$ is a {\em transformed} policy that takes advantage of the availability of information about the unobserved variables summarized in $(\bpi_t)_{t\in\MT}$.  In particular, since in APOMDPs \citep{Saghafian2018} we have access to $(\bpi^m_t)_{t\in\MT}$ under each model $m\in\MM$, we might be able to make use of it to ensure that $\bmu^b$ satisfies sequential ignorability. For example, in many medical decision-making applications, including  the case study with our partner hospital discussed in Section \ref{sec:case}, the immediate gain  depends on hidden states (confounders) only through $(\bpi^m_t)_{t\in\MT}$. This occurs, for example, when the gain in each period $t$ is simply $(\bpi^m_t)'\,\g^a_t$, where $\g^a_t\teq (g(s_t, a_t))_{s_t\in\MS}$; that is, when the gain in each period $t$ is considered to be the average value of $\g^a_t$ considering the belief distribution $\bpi^m_t$.\footnote{This is the case in how the value function in traditional POMDPs is often constructed; see, e.g., Eq. \eqref{traditional}.} (See  \cite{Boloori2020} for more discussions about the immediate gain values in our medical decision-making application, and \cite{Xu2020} for a similar assumption made in analyzing latent-state models.) However, $\bmu^b$ might not satisfy sequential ignorability in some other applications; see, e.g., \cite{Saghafian2021} for related challenges in mHealth applications. Thus, in Section \ref{sec:extension}, we provide an extension of Proposition \ref{prop:POMDPest} by making use of  the BUC results presented in Proposition \ref{prop:buc}. However, in what follows, we first focus on understanding and analyzing the cases where $\bmu^b$ satisfies sequential ignorability, as it provides a foundation for  the extensions discussed in Section \ref{sec:extension}.

\end{remark}

\begin{remark} [\textbf{Effective Approximation}] In using the results of Proposition \ref{prop:POMDPest},  we only require an {\em approximate} solution to \eqref{eq:POMDPest3}. Thus, how the solution to \eqref{eq:POMDPest3} is obtained is not that restrictive. Indeed, there are many ways to obtain an approximate solution to \eqref{eq:POMDPest3}. In what follows, however, we provide an effective way of estimating the optimal policy and optimal value function using \eqref{eq:POMDPest3}. We do so by taking advantage of  important structural properties of the optimal value function of POMDPs and APOMDPs. Specifically, the optimal value function of POMDPs is known to be piecewise linear and convex in $\bpi$ under some mild conditions \citep{Sondik_Finite_1973}. \cite{Saghafian2018} shows that in general the convexity does not hold in APOMDPs, and some additional conditions are needed (see Proposition 2 of \cite{Saghafian2018}). To be consistent, for both POMDP and APOMDP settings, we only assume {\em piecewise linearity} and {\em continuity}, but do not impose any assumption on convexity. This, in turn, helps us in another way: while piecewise linear and continuous functions can be effectively learned from data, learning a function that is both piecewise linear and convex (i.e., is {\em point-wise maximum} of a set of linear functions) is much harder \citep[see, e.g.,][and the references therein]{Magnani2009}.
\end{remark}

Let $\MV$ denote the set of  real-valued piecewise linear and continuous bounded functions defined on $\Delta_\MS$, and assume $V_t^{m,\bmu^e}\in\MV$. To learn $V_t^{m,\bmu^e}\in\MV$ using \eqref{eq:POMDPest3}, we consider the parametric version of the value function:  $V_t^{m,\bmu^e}(\bpi;\bpsi_t)\teq \big(\bb (\bpi))'\, \bpsi_t$, where $\bb (\bpi)\teq\big(\bb_1 (\bpi), \bb_2 (\bpi),\cdots, \bb_{d_t} (\bpi)\big)'$ is a predefined {\em basis function} that allows us to ensure that the learned function is in $\MV$, and $\bpsi_t \in \bPsi_t\subseteq \mathbb R^{d_t}$ is the parameter.\footnote{Allowing the dimensionality of the parameter space, $d_t$, to depend on $t$ can enable us increase flexibility as $t$ grows (e.g., by introducing more knots). The special case where $d_t$ does not depend on $t$ is still useful in some settings, including those where the goal is to learn the long-run impact of a policy (see, e.g., Algorithms \ref{alg:DAV} and \ref{alg:SAV} in the next sections).} This also enables us to set $\phi(\bpi)\teq \bb (\bpi)$ in \eqref{eq:POMDPest3},  since $\bb (\bpi)$ can be thought of as the gradient of $V_t^{m,\bmu^e}(\bpi;\bpsi_t)$ with respect to its parameter, which only depends on $\bpi$ (and not the parameter) and is almost everywhere defined.

Furthermore, since $\bpsi_{t}$ can be high-dimensional in some applications (especially when $t$ is large), we estimate it using a regularized approach as follows (to avoid overfitting). Starting with $V_0^m(\bpi)=0$ and moving backwards iteratively, having an estimation of $T-t$ periods to go value function in hand ($\hat V_{T-t}^{m,\bmu^e}$), we define
\begin{equation} \label{eq:POMDPest4}
\varphi^{m,\bmu^e} (\bpsi_t)\teq \E^{\mathbb P} \bigg[ \frac{\mu_t^e(A_t|\bPi^m_t)}{\mu_t^b(A_t|\bPi^m_t)}\Big[G_t+ \beta\, \hat V_{T-t}^{m,\bmu^e}  (T(\bPi^m_t, A_t, O_t, m))- V_{T-t+1}^{m,\bmu^e}  (\bPi^m_t; \bpsi_t)\Big] \bb(\bPi^m_t)\bigg].
\end{equation}
We then obtain the estimate
\begin{equation} \label{eq:POMDPest5}
\hat \bpsi_t^{\bmu^e}=\arg\!\min_{\bpsi_t\in\bPsi_t} \bigg\{\big(\varphi^{m,\bmu^e} (\bpsi_t)\big)'\, \bOmega\, \varphi^{m,\bmu^e} (\bpsi_t)+ \theta_t \mathcal P (\bpsi_t)\bigg\},
\end{equation}
where $\bOmega$ is an arbitrary  positive definite matrix, $\mathcal P (\cdot)$ is a penalty function, and $\theta_t$ is a tuning parameter.\footnote{In our case study, simulations experiments, and theoretical results, we make use of the squared Euclidean norm as the penalty function, and hence, assume $\mathcal P (\bpsi_t)= \bpsi_t'\bpsi_t$.}  Consequently, we plug in $\hat \bpsi_t^{\bmu^e}$ in $V_{T-t+1}^{m,\bmu^e}  (\bpi_t; \bpsi_t)$ and thereby obtain an estimate for the value function $V_{T-t+1}^{m,\bmu^e}  (\bpi_t)$, and move to the next period (backward).  This procedure, under a given model $m\in\MM$, yields an estimator for the gain under $\bmu^e$. That is, $\hat \Gamma_T^m (\bmu^e)\teq \int \hat V_{T}^{m,\bmu^e} (\bpi)\, dF(\bpi)$ can be used as an estimator for $\Gamma_T^m (\bmu^e)= \int  V_{T}^{m,\bmu^e} (\bpi)\, dF(\bpi)$, where $dF(\bpi)$ is a given distribution on (starting) belief values. Since we have an estimator for the gain under any policy $\bmu^e$, we can obtain $\hat\bmu{^{e*}}\teq\arg\!\max_{\bmu^e\in\Upsilon} \hat \Gamma_T^m (\bmu^e)$ as an estimate for the optimal policy under model $m$, where $\Upsilon$ is  a given set of policies.\footnote{We use the ``$\max$" operator instead of ``$\sup$," because in most real-world applications, $\Upsilon$ is first identified by a set of domain experts and is such that maximum is obtained. We later make this assumption more implicit (see, e.g., Condition (C4) in Section \ref{sec:assymptotics}. Furthermore, in various practical applications, $\Upsilon$ is often restricted to the set of policies that satisfy specific attributes such as fairness or interpretability.}  Finally, an estimate of the gain under the optimal policy is $\hat \Gamma_T^m (\hat\bmu{^{e*}})$.

In an infinite-horizon setting, the procedure above simplifies. This is because in homogenous POMDPs (and APOMDPs) the value function with $t$ periods to go converges to a stationary value function as  $t\to\infty$ \citep[see, e.g., Proposition 1 of][]{Saghafian2018}. Therefore, in \eqref{eq:POMDPest3} we can replace both $V^{m,\bmu^e}_{T-t} (\cdot)$ and $V^{m,\bmu^e}_{T-t+1} (\cdot)$ with the same function. This removes the need for recursive calculations and allows us to follow a ``one-shot" method. We discuss this further in the next sections, and also  study the asymptotic behavior of our proposed approach.

\subsection{Augmented V-Learning for APOMDPs}
Motivated by the results in the previous section, we now extend our approach to APOMDPs, where the condition $|\MM|=1$ does not hold.  We propose two approaches termed {\em Direct Augmented V-Learning} ($\DAV$) and {\em Safe Augmented V-Learning} ($\SAV$).  As we will see, in $\DAV$, we {\em directly} extend the approach presented in the previous section for POMDPs by first obtaining  a value function separately for each POMDP model in $\MM$. These values are then combined at the end of the horizon to provide an estimate of the value function for the APOMDP. In $\SAV$, however, we make use of a {\em safe} estimation  approach upfront that takes into account ambiguity and removes the need to obtain  a value function separately for each POMDP model in $\MM$. 

\subsubsection{Direct Augmented V-Learning ($\DAV$).}\label{sec:DAV}
Recall that for each evaluation policy $\bmu^e$ and each given $m\in\MM$, we can use the approach proposed for POMDPs  in Section \ref{sec:AVPOMDP} to obtain an estimate for the value function $V_{T}^{m,\bmu^e}  (\cdot)$, which we denote by $\hat V_{T}^{m,\bmu^e}  (\cdot)$. Thus,  we can first obtain an estimate for the APOMDP value function:
\begin{equation}
\hat V_{T}^{\bmu^e} (\bpi) = MEU_\alpha \big[\hat V_{T}^{m,\bmu^e} (\bpi)\big] \teq \alpha\,\inf_{m\in\MM} \hat V_{T}^{m,\bmu^e} (\bpi)+ (1-\alpha)\,\sup_{m\in\MM} \hat V_{T}^{m,\bmu^e}(\bpi).
\end{equation}
Next, to estimate the optimal policy, we note that for any policy $\bmu^e$, the estimator of the gain is $\hat \Gamma_T ({\bmu^e})= \int \hat V_{T}^{\bmu^e} (\bpi)\, dF(\bpi)$,
where $dF(\bpi)$ is a given distribution on (starting) belief values. This means that we
 can obtain an estimate of the optimal  policy  as $\hat \bmu{^{e*}}\teq\arg\!\max_{\bmu^e\in\Upsilon} \hat \Gamma_T ({\bmu^e})$. Finally, the estimated optimal gain is $\hat \Gamma_T (\hat \bmu{^{e*}})$.

This $\DAV$ approach for APOMDPs in the infinite-horizon case is presented in Algorithm \ref{alg:DAV}. In presenting this algorithm, as is often the case, we assume that the data  only includes a finite number of periods for each subject, but the goal is to estimate the long-run performance of  policies \citep[see, e.g.,][]{Luckett2020, Xu2020}.  We also use subscript $n$ to highlight the dependency of  our estimators to the number of trajectories in the data set, which in turn allows us to investigate the behavior of our  proposed learning algorithm  as $n\to\infty$ (see Section \ref{sec:assymptotics}). Our estimation equations for the infinite-horizon gain are
\begin{equation} \label{eq:SAV4}
\varphi_n^{m,\bmu^e} (\bpsi)\teq \E^{\mathbb P} \Bigg[\sum_{t\in\MT}\bigg[ \frac{\mu^e(A_t|\bPi^m_t)}{\mu^b(A_t|\bPi^m_t)}\Big[G_t+ \beta\,  V_\infty^{m,\bmu^e}  (T(\bPi^m_t, A_t, O_t, m))- V_\infty^{m,\bmu^e} (\bPi^m_t)\Big] \bb (\bPi^m_t)\bigg]\Bigg]
\end{equation}
and
\begin{equation} \label{eq:SAV5}
\hat \bpsi_n^{m,\bmu^e}=\arg\!\min_{\bpsi\in\bPsi} \bigg\{\big(\varphi_n^{m,\bmu^e} (\bpsi)\big)'\, \bOmega\, \varphi_n^{m,\bmu^e} (\bpsi)+ \theta_n \mathcal P (\bpsi)\bigg\},
\end{equation}
where  $\bPsi\subseteq \mathbb R^{d}$. Similar to before, we make use of the piecewise linearity and continuity  of the value function (i.e., the fact that $V^{m,\bmu^e}_{\infty}\in \MV)$ for all $m\in\MM$. This  allows us to use predefined basis function to ensure that the learned function remains in $\MV$ when we use the parametric form $ V^{m,\bmu^e}_{\infty} (\bpi,\bpsi)\teq \big(\bb (\bpi))'\, \bpsi$.

  Using \eqref{eq:SAV5}, we then set $\hat V^{m,\bmu^e}_{\infty}  (\bpi)\teq V^{m,\bmu^e}_{\infty}  (\bpi; \hat \bpsi_n^{m,\bmu^e})$. In addition, denoting the infinite-horizon gain under any policy  $\bmu^e$  and $m\in\MM$ by $\Gamma^m_{\infty} (\bmu^e)\teq \int  V_{\infty}^{m,\bmu^e} (\bpi)\, dF(\bpi)$, we consider $\hat \Gamma^m_{\infty} (\bmu^e)\teq \int  \hat V_{\infty}^{m,\bmu^e} (\bpi)\, dF(\bpi)$ as an estimator for $\Gamma^m_{\infty} (\bmu^e)$. With estimated values under each model $m$ in hand, we next define the estimated overall gain (a model independent value) as $\hat \Gamma_{\infty} (\bmu^e)\teq \alpha \inf_{m\in\MM} \hat \Gamma^m_{\infty} (\bmu^e) + (1-\alpha) \sup_{m\in\MM} \hat \Gamma^m_{\infty} (\bmu^e)$, which provides an estimation for the overall gain $\Gamma_{\infty} (\bmu^e)\teq \alpha \inf_{m\in\MM} \Gamma^m_{\infty} (\bmu^e) + (1-\alpha) \sup_{m\in\MM} \Gamma^m_{\infty} (\bmu^e)$.

Finally, the estimated optimal policy and its infinite-horizon value for the APOMDP are obtained as $\hat\bmu{^{e*}}\teq \arg\!\max_{\bmu^e\in\Upsilon} \hat \Gamma_{\infty} (\bmu^e)$ and  $\hat \Gamma_{\infty} (\hat\bmu^{e*})=\max_{\bmu^e\in\Upsilon} \hat \Gamma_{\infty} (\bmu^e)$, respectively, where the latter  provides an estimate for $\Gamma_{\infty} (\bmu^{e*})\teq \max_{\bmu^e\in\Upsilon} \Gamma_{\infty} (\bmu^{e*})$. Similarly, under each model $m$, we denote the estimated optimal policy and its infinite-horizon value as $\hat\bmu{^{e*,m}}\teq \arg\!\max_{\bmu^e\in\Upsilon} \hat \Gamma^m_{\infty} (\bmu^e)$, and  $\hat \Gamma^m_{\infty} (\hat\bmu^{e*,m})=\max_{\bmu^e\in\Upsilon} \hat \Gamma^m_{\infty} (\bmu^e)$, respectively, where the latter  provides an estimate for $\Gamma^m_{\infty} (\bmu^{e*,m})\teq \max_{\bmu^e\in\Upsilon} \Gamma^m_{\infty} (\bmu^{e*,m})$.

\begin{algorithm}[t]\label{alg:DAV}
\scriptsize
\DontPrintSemicolon
   \For{each observed trajectory  and model $m\in\MM$}
        {
        	Initialize  $\bpi^m_0$ using a random draw from  $F(\bpi)$;  \;
            set t=1;\;
             \While{$t+1\in\MT$}
               {
                 $\bpi^m_{t+1}\leftarrow T(\bpi^m_t, a_t, o_t, m)$;
               }

        }

 \For{any given $\bmu^e\in\Upsilon$ and $m\in\MM$}
 {
  $ \varphi_n^{m,\bmu^e} (\bpsi)\leftarrow \E^{\mathbb P} \Bigg[\sum_{t\in\MT}\bigg[ \frac{\mu^e(A_t|\bPi^m_t)}{\mu^b(A_t|\bPi^m_t)}\Big[G_t+ \beta\,  V_\infty^{m,\bmu^e}  (T(\bPi^m_t, A_t, O_t, m))- V_\infty^{m,\bmu^e} (\bPi^m_t)\Big] \bb (\bPi^m_t)\bigg]\Bigg]$; \;

$\hat \bpsi_n^{m,\bmu^e}\leftarrow\arg\!\min_{\bpsi\in\bPsi} \bigg\{\big(\varphi_n^{m,\bmu^e} (\bpsi)\big)'\, \bOmega\, \varphi_n^{m,\bmu^e} (\bpsi)+ \theta_n \mathcal P (\bpsi)\bigg\}$;\;

$\hat V^{m,\bmu^e}_{\infty}(\bpi)  \leftarrow \big(\bb (\bpi))'\, \hat \bpsi_n^{m,\bmu^e}$;\;

$\hat \Gamma^m_{\infty} (\bmu^e) \leftarrow\int  \hat V_{\infty}^{m,\bmu^e} (\bpi)\, dF(\bpi)$;\;

}

 \For{any given $\bmu^e\in\Upsilon$}
 {
$\hat \Gamma_{\infty} (\bmu^e) \leftarrow \alpha \inf_{m\in\MM} \hat \Gamma^m_{\infty} (\bmu^e) + (1-\alpha) \sup_{m\in\MM} \hat \Gamma^m_{\infty} (\bmu^e)$;\;

}

$\hat\bmu{^{e*}}  \leftarrow \arg\!\max_{\bmu^e\in\Upsilon} \hat \Gamma_{\infty} (\bmu^e)$;\;

$\hat \Gamma_{\infty} (\hat\bmu^{e*} )\leftarrow \max_{\bmu^e\in\Upsilon} \hat \Gamma_{\infty} (\bmu^e)$;

\caption{$\DAV$}
\end{algorithm}

\subsubsection{Safe Augmented V-Learning ($\SAV$).} \label{sec:SAV}
The $\DAV$ algorithm presented in the previous section is a direct extension of the approach proposed for POMDPs (Section \ref{sec:AVPOMDP}) in which ``the curse of ambiguity" \citep{Saghafian2018} is overcome at the end. In contrast, in $\SAV$, this curse is overcome upfront via  a ``safe method" for estimating the underlying parameter $\bpsi_t$, and hence, the value function. To develop the $\SAV$ algorithm, similar to before, we first  denote the APOMDP value function with $t$ periods to go under policy $\bmu^e$ (a model independent function) with $V^{\bmu^e}_t$, assume that $V^{\bmu^e}_t\in\MV$,  and parameterize it via $V_t^{\bmu^e}(\bpi;\bpsi_t)\teq \big(\bb (\bpi)\big)'\, \bpsi_t$. We then estimate its parameter as

\begin{equation} \label{eq:SAV2}
\hat \bpsi_t^{\bmu^e} \teq MEU_{\alpha}\big[\hat \bpsi_t^{m,\bmu^e}\big] \teq \alpha \, \hat \bpsi_t^{\un m,\bmu^e}+ (1-\alpha)\, \hat\bpsi_t^{\ov m,\bmu^e},
\end{equation}
where $\alpha\in\MI$ can be viewed as a tuning parameter, $\un m\teq \arg\!\inf_{m\in\MM} ||\hat\bpsi_t^{m,\bmu^e}||$, $\ov m\teq \arg\!\sup_{m\in\MM} ||\hat\bpsi_t^{m,\bmu^e}||$,\footnote{We assume $\MM$ is such that $\inf_{m\in\MM} ||\bpsi_t^{m,\bmu^e}||$ and $\sup_{m\in\MM} ||\bpsi_t^{m,\bmu^e}||$ are both finite, and $\un m$ and $\ov m$ are both in $\MM$.} and
\begin{equation} \label{eq:SAV3}
\hat \bpsi_t^{m,\bmu^e}=\arg\!\min_{\bpsi_t\in\bPsi_t} \bigg\{\big(\varphi^{m,\bmu^e} (\bpsi_t)\big)'\, \bOmega\, \varphi^{m,\bmu^e} (\bpsi_t)+ \theta_t \mathcal P (\bpsi_t)\bigg\},
\end{equation}
where $\varphi^{m,\bmu^e} (\bpsi_t)$ is defined in \eqref{eq:POMDPest4}.  Consequently, we plug  $\hat \bpsi_t^{\bmu^e}$ obtained in \eqref{eq:SAV2} in $V_{T-t+1}^{\bmu^e}  (\bpi_t; \bpsi_t)$, which yields an estimate for the APOMDP value function $V_{T-t+1}^{\bmu^e}  (\bpi_t)$, and move to the next period (backwards) as before. This yields
an estimated value function $\hat V_{T}^{\bmu^e}  (\bpi)$. Denoting the gain under any policy  $\bmu^e$ by $\Gamma_T (\bmu^e)\teq \int  V_{T}^{\bmu^e} (\bpi)\, dF(\bpi)$, we use $\hat \Gamma_T (\bmu^e)\teq \int  \hat V_{T}^{\bmu^e} (\bpi)\, dF(\bpi)$ as an estimator for $\Gamma_T (\bmu^e)$.

Finally, optimization over $\bmu^e\in\Upsilon$ will provide the estimated optimal policy of the APOMDP under the $\SAV$ approach: $\hat\bmu{^{e*}}\teq \arg\!\max_{\bmu^e\in\Upsilon} \hat \Gamma_T (\bmu^e)= \arg\!\max_{\bmu^e\in\Upsilon} \int \hat V_{T}^{\bmu^e} (\bpi)\, dF(\bpi)$. The estimated optimal gain under this approach is $\hat \Gamma_T (\hat\bmu^{e*})=\max_{\bmu^e\in\Upsilon} \int \hat V_{T}^{\bmu^e} (\bpi)\, dF(\bpi)$, which provides an estimate for $\Gamma_T (\bmu^{e*})\teq \max_{\bmu^e\in\Upsilon} \int  V_{T}^{\bmu^e} (\bpi)\, dF(\bpi)$. Similar to before, this procedure can also be used for the infinite-horizon case by noting that since both $V_{T-t}(\cdot)$ and $V_{T-t+1}(\cdot)$  become $V_\infty(\cdot)$ the calculations simplifies. The $\SAV$ approach for infinite-horizon case is presented in Algorithm \ref{alg:SAV}. Besides their benefit in analyzing the long-run impact of  different treatment regimes, both Algorithms \ref{alg:DAV} and  \ref{alg:SAV}  can also be used as  {\em approximations} for learning policies that work well over a finite but long horizon.

\begin{algorithm}[t]\label{alg:SAV}
\scriptsize
\DontPrintSemicolon
   \For{each observed trajectory  and model $m\in\MM$}
        {
        	Initialize  $\bpi^m_0$ using a random draw from  $F(\bpi)$;  \;
            set t=1;\;
             \While{$t+1\in\MT$}
               {
                 $\bpi^m_{t+1}\leftarrow T(\bpi^m_t, a_t, o_t, m)$;
               }

        }

 \For{any given $\bmu^e\in\Upsilon$ and $m\in\MM$}
 {
  $ \varphi_n^{m,\bmu^e} (\bpsi)\leftarrow \E^{\mathbb P} \Bigg[\sum_{t\in\MT}\bigg[ \frac{\mu^e(A_t|\bPi^m_t)}{\mu^b(A_t|\bPi^m_t)}\Big[G_t+ \beta\,  V_\infty^{m,\bmu^e}  (T(\bPi^m_t, A_t, O_t, m))- V_\infty^{m,\bmu^e} (\bPi^m_t)\Big] \bb (\bPi^m_t)\bigg]\Bigg]$; \;

$\hat \bpsi_n^{m,\bmu^e}\leftarrow\arg\!\min_{\bpsi\in\bPsi} \bigg\{\big(\varphi_n^{m,\bmu^e} (\bpsi)\big)'\, \bOmega\, \varphi_n^{m,\bmu^e} (\bpsi)+ \theta_n \mathcal P (\bpsi)\bigg\}$;\;

}

 \For{any given $\bmu^e\in\Upsilon$}
 {

$\un m \leftarrow \arg\!\inf_{m\in\MM} ||\hat\bpsi_n^{m,\bmu^e}||$; \;

$\ov m \leftarrow \arg\!\sup_{m\in\MM} ||\hat\bpsi_n^{m,\bmu^e}||$; \;

$\hat \bpsi^{\bmu^e}_n \leftarrow \alpha \, \hat \bpsi_n^{\un m,\bmu^e}+ (1-\alpha)\, \hat\bpsi_n^{\ov m,\bmu^e}$; \;

$\hat V^{\bmu^e}_{\infty}(\bpi)  \leftarrow \big(\bb (\bpi))'\, \hat \bpsi_n^{\bmu^e}$;\;

$\hat \Gamma_{\infty} (\bmu^e) \leftarrow \int  \hat V_{\infty}^{\bmu^e} (\bpi)\, dF(\bpi)$;\;

}
$\hat\bmu{^{e*}}  \leftarrow \arg\!\max_{\bmu^e\in\Upsilon} \hat \Gamma_{\infty} (\bmu^e)$;\;

$\hat \Gamma_{\infty} (\hat\bmu^{e*} )\leftarrow \max_{\bmu^e\in\Upsilon} \hat \Gamma_{\infty} (\bmu^e)$;

\caption{$\SAV$}
\end{algorithm}

\section{Performance Analyses: Theoretical Results}\label{sec:assymptotics}
We now establish some theoretical results for the performance of our proposed approaches. Specifically, we demonstrate the asymptotic properties of the estimators under our  main proposed algorithm, $\DAV$ (Algorithm \ref{alg:DAV}). With some minor modifications, one can then also establish similar results for the estimators under the second proposed approach, $\SAV$ (Algorithm~\ref{alg:SAV}).\footnote{For general results related to the asymptotic properties of V-Learning algorithms when all variables are observable and there is no model ambiguity, we refer interested readers to \cite{Luckett2020}.}

The main results of this section are as follows. Under some conditions discussed below, we first establish  weak consistency and asymptotic normality of the estimators under any policy  $\bmu^e\in\Upsilon$ (Theorem~\ref{theo:asymptotics:fixed}). We then move to the estimators related to the optimal policy, and  establish weak consistency and asymptotic normality of both the estimated optimal policy and its estimated value (Theorem~\ref{theo:asymptotics:opt}). To establish our results, we make use of arguments in {\em empirical processes} (specifically for stationary process as opposed to i.i.d. ones; see, e.g., \cite{Dedecker2002, Kosorok2008}), and think of each realization of the underlying stochastic process as a function in $\ell^\infty(\Upsilon)$ (i.e., the set of real-valued bounded functions indexed by $\bmu^e\in\Upsilon$).

We assume $\bOmega$ in \eqref{eq:SAV5} is an arbitrary  positive-definite matrix, $\mathcal P (\cdot)$ is the squared norm penalty function, and $\theta_n$ is a tuning parameter satisfying $\theta_n=o_p(n^{-1/2})$. We also assume that $\E^m\big[||\bb(\bPi_t)||^2\big]$ and $\E^m \big[G_t^2\big]$ are both finite for all $m\in\MM$ and $t\in\MT$.   Some other technical conditions are needed, mainly because of two broad set of challenges in our setting which make establishing asymptotic results more involved:  (1) the underlying process is not i.i.d over time, and (2) there is model ambiguity ($|\MM|\neq 1$).  Specifically,  we need the following ``regularity" conditions on the parameter space, trajectories space, policy space, and model space:

\begin{itemize}[leftmargin=1cm,align=left]
\item[(C1)]  For every $\bmu^e\in\Upsilon$ and $m\in\MM$ there exists a unique solution to $\varphi^{m,\bmu^e} (\bpsi)=0$ denoted by $\bpsi_\diamond^{m,\bmu^e}\in\bPsi\subseteq \mathbb R^d$, where $\sup_{\bmu^e\in\Upsilon}||\bpsi_\diamond^{m,\bmu^e}||<\infty$,  $\bpsi_\diamond^{m,\bmu^e}$  is an interior point of $\bPsi$, and $\bPsi$ is compact subset of $\mathbb R^d$.
\item[(C2)]  There exists a  $2<\rho<\infty$ such that for all $m\in\MM$:
\begin{itemize}[leftmargin=1.5cm]
\item[(C2a)] The class of policies ($\Upsilon$) is either finite, or its bracketing integral satisfies  $J_{[]}(\infty,\Upsilon, L_\rho(P^m))<\infty$, where $P^m$ is  the marginal stationary distribution of the sequence $\{(\bPi^m_t, A_t)\}_{t\geq 1}$.\footnote{For the definition of the bracketing integral, $J_{[]}(\infty,\Upsilon, L_\rho(P^m))$, see, e.g., \cite{Kosorok2008}.}
\item[(C2b)] The sequence $\{(\bPi^m_t, A_t)\}_{t\geq 1}$  is an absolutely regular stationary process with its $\beta$-mixing coefficients $\zeta^m (t)$ satisfying $\sum_{t=1}^{\infty} k^{2/(\rho-2)} \zeta^m (t)<\infty$.\footnote{For the definition of  an absolutely regular stationary process and its $\beta$-mixing coefficients, see, e.g., \cite{Dedecker2002, Kosorok2008}, and the references therein.}
\end{itemize}
\noindent\item[(C3)] There exists a constant $c_1>0$ such that for all $m\in\MM$, $t\in\MT$, $\bmu^e\in\Upsilon$, and  $\mathbf{c}\in\mathbb R^d$:
\begin{equation}
\mathbf{c}'\,\E^{m} \Bigg[\frac{\mu^e(A_t|\bPi^m_t)}{\mu^b(A_t|\bPi^m_t)}\, \bb(\bPi_t^m)\, \Big(\bb(\bPi_t^m)-\beta\, \bb(T(\bPi^m_t, A_t, O_t, m))\Big)'\Bigg]
 \mathbf{c}\geq c_1 ||\mathbf{c}||^2.
\end{equation}
\noindent\item[(C4)] $\bmu^{e*}$ is a unique and well septated maximizer of $\Gamma_\infty(\bmu^{e})$ and $\bmu^{e*}$ is in the interior $\Upsilon$.
\noindent\item[(C5)] For every $\bmu^e\in\Upsilon$: $|\inf_{m\in\MM} \Gamma^m_{\infty} (\bmu^e)|<\infty$, $|\sup_{m\in\MM} \Gamma^m_{\infty} (\bmu^e)|<\infty$, and  $\MM$ contains both $\arg\!\inf_{m\in\MM} \Gamma^m_{\infty} (\bmu^e)$ and $\arg\!\sup_{m\in\MM} \Gamma^m_{\infty} (\bmu^e).$
\end{itemize}

Assumptions related to these conditions  are relatively common in  the Z-estimation and M-estimation theories \citep[see, e.g.,][]{Kosorok2008}. Some of these conditions are also assumed to hold in the {\em Generalized Method of Moments} (GMM)  \citep[for asymptotic properties of GMM, see, e.g.,][]{Hansen1982}. These conditions hold both in our case study of NODAT patients (Section \ref{sec:case}) and in our simulation experiments (Section \ref{sec:synthetic}).  (C1) is a regularity condition on the parameter space, and ensures that the solutions obtained by solving $\varphi^{m,\bmu^e} (\bpsi)=0$ are ``well-behaved." (C2a) is a regularity condition on the policy space, and  requires that the set of policies under consideration satisfy a minimum level of ``complexity" (measured by an appropriate {\em entropy-based} metric). This condition clearly allows working with any finite set of policies, but also holds for many infinite sets of policies \citep[see, e.g., the parametric class of policies in][]{Luckett2020}.    (C2b) is a regulatory condition on the space of trajectories and allows viewing their formation as a suitable stationary process. The $\beta$-mixing coefficients $\zeta^m (t)$ quantify dependency of the observed values in the process $t$ steps removed, and are zero when there is no such dependency. (C3) ensures that the matrix $\bC^m(\bmu^e)$ defined in Theorem \ref{theo:asymptotics:fixed} below is positive-definite, and hence, invertible. One can empirically check whether (C3) holds by creating certain matrixes using data and testing whether they are positive-definite. (C4) is needed to establish that the sequence of estimated optimal policies converges to the true optimal policy, which is a stronger result than just the gain of these policies converging to each other. (C5) is a regularity condition on the space of models, $\MM$, which holds in most real-wrold applications, because any set of models can be represented/approximated with a finite set (with any required level of accuracy).

We first establish the asymptotic behavior of our estimators under any given policy  $\bmu^e\in\Upsilon$ by only requiring (C1)-(C3). The proof  is based on some additional results provided in Appendix B (see Lemmas \ref{ec: lemma:asymptotics:Donsker} and \ref{ec: lemma:asymptotics:11.24}), which establish Donsker properties and asymptotic normality in $\ell^\infty (\Upsilon)$ for the underlying absolutely regular stationary process in our setting .

\begin{theorem}[Asymptotic Behavior: Fixed Policy and its Value]\label{theo:asymptotics:fixed} Suppose (C1)-(C3) hold and the behavior policy satisfies positivity. Then under $\DAV$ (Algorithm \ref{alg:DAV}), for any  $\bmu^e\in\Upsilon$ and $m\in\MM$, we have:
\begin{itemize}[leftmargin=1cm,align=left]
\item[(i)]  $\hat \bpsi_n^{m,\bmu^e}\overset{p}{\to} \bpsi_\diamond^{m,\bmu^e}$.
\item[(ii)] $\sqrt{n} \,\big[\hat \bpsi_n^{m,\bmu^e}- \bpsi_\diamond^{m,\bmu^e}\big]\overset{d}{\to} \mathbb{G}(\bmu)$ in $\ell^\infty (\Upsilon)$, where $\mathbb{G}(\bmu)$ is a zero-mean and tight Gaussian process indexed by $\bmu\in\Upsilon$ with the covariance function given by
\begin{equation}
\E\Big[\mathbb{G}(\bmu)\mathbb{G}(\tilde\bmu)\Big]= \Big(\bC^m(\bmu^e)\Big)^{-1} \, \tilde\bC^m(\bmu^e, \tilde\bmu^e)\, \Big(\big(\bC^m(\bmu^e)\big)^{-1}\Big)' \ \ \  \ \ \forall\bmu,\tilde{\bmu}\in\Upsilon,
\end{equation}
where
\begin{equation}
\bC^m(\bmu^e)\teq \E^{m} \Bigg[\frac{\mu^e(A_t|\bPi^m_t)}{\mu^b(A_t|\bPi^m_t)}\, \bb(\bPi_t^m)\, \Big(\bb(\bPi_t^m)-\beta\, \bb(T(\bPi^m_t, A_t, O_t, m))\Big)'\Bigg],
\end{equation}
 \begin{equation}
\tilde\bC^m(\bmu^e, \tilde\bmu^e)\teq \E^{m} \Bigg[\frac{\mu^e(A_t|\bPi^m_t)\tilde\mu^e(A_t|\bPi^m_t)}{\mu^b(A_t|\bPi^m_t)\,\mu^b(A_t|\bPi^m_t)}\, \bvartheta(\bPi^m_t,\bpsi_\diamond^{\bmu^e})\, \bvartheta(\bPi^m_t,\bpsi_\diamond^{\tilde\bmu^e})\,\bb(\bPi_t^m)\, \Big(\bb(\bPi_t^m)\Big)' \Bigg],
\end{equation}
and \begin{equation}
\bvartheta(\bPi^m_t,\bpsi_\diamond^{\bmu^e})\teq G_t+ \Big[\beta\, \bb(T(\bPi^m_t, A_t, O_t, m))-\bb(\bPi_t^m)\Big]\bpsi_\diamond^{\bmu^e}.
\end{equation}
\item[(iii)] $\hat \Gamma^m_\infty (\bmu^e) \overset{p}{\to} \Gamma^m_\infty (\bmu^e)$.
\item[(iv)]  $\hat \Gamma_\infty (\bmu^e) \overset{p}{\to} \Gamma_\infty (\bmu^e)$ assuming (C5) holds.
\end{itemize}
\end{theorem}

We next establish the asymptotic properties of the optimal policy and the gain under it. The proof of the following theorem is based on an additional result provided in Appendix B (see Lemma \ref{ec: lemma:asymptotics:M-Estimation}), which in turn relies on results from the $M$-estimation theory.

\begin{theorem}[Asymptotic Behavior: Optimal Policy and its Value]\label{theo:asymptotics:opt} Suppose (C1)-(C5) hold and the behavior policy satisfies positivity. Then, considering a metric space $(\Upsilon, d_{\Upsilon})$, under $\DAV$ (Algorithm \ref{alg:DAV}) we have:
\begin{itemize}[leftmargin=1cm,align=left]
\item[(i)] $d_{\Upsilon}(\hat\bmu^{e*,m},\bmu^{e*,m})\overset{p}{\to} 0$ for all $m\in\MM$.
\item[(ii)] $d_{\Upsilon}(\hat\bmu^{e*},\bmu^{e*})\overset{p}{\to} 0$.
\item[(ii)] $\hat \Gamma^m_{\infty}(\hat \bmu^{e*,m}) \overset{p}{\to} \Gamma^m_{\infty}(\bmu^{e*,m})$.
\item[(iv)] $\hat \Gamma_{\infty}(\hat \bmu^{e*}) \overset{p}{\to} \Gamma_{\infty}(\bmu^{e*})$.
\end{itemize}
\end{theorem}

\section{Extension: Learning Under Bounded Unobservable Confounding (BUC)}\label{sec:extension}
As discussed in Remark \ref{remark:BUC}, in various applications, one might be able to ensure that $\bmu^b$ satisfies sequential ignorability, because it takes advantage of the availability of information about the unobserved variables (confounders) summarized in $(\bpi_t)_{t\in\MT}$. The $\DAV$ and $\SAV$ approaches introduced earlier are based on the results of Proposition \ref{prop:POMDPest}, which assumes $\bmu^b$ satisfies sequential ignorability. In this section, we show how such results can be extended to cases where $\bmu^b$ does not satisfy sequential ignorability, but satisfies BUC conditions introduced earlier (see, e.g., Definition~\ref{def:boundedconf}). This, in turn, allows us to extend $\DAV$ and $\SAV$, and introduce their BUC counterparts, which we term  $\DAVBUC$ and $\SAVBUC$, respectively.

Similar to \eqref{eq:BUCEquivalent}, which is based on Lemma \ref{lemma:ec:boundedconf} (Appendix B),  assume for each model $m\in\MM$ there exist constants $\eta^m_t\in[1,\,\infty)$  such that:
\begin{equation}\label{eq:BUCEquivalent2}
(\eta^m_t)^{-1}\leq\frac{\mu^b_t (a_t|\bPi_t^m, \bS_t^m)}{\mu^b_t (a_t|\bPi_t^m)}\leq \eta^m_t\ \ \ \ \ a.s.
\end{equation}
over $\bPi_t^m$ and $\bS_t^m$ for all $t\in\MT$ and $a\in\MA$. If $\eta^m_t=1$, $\mu^b_t$ satisfies sequential ignorability. Furthermore, with $\eta^m_t=1$, \eqref{eq:BUCEquivalent2} implies that benefiting from  $\bPi_t^m$ under each model $m$ and making use of marginalized treatment propensities $\mu^b_t (a_t|\bPi_t^m)$  is enough for the goal of estimating the true treatment propensities $\mu^b_t (a_t|\bPi_t^m, \bS_t^m)$. More broadly, however, \eqref{eq:BUCEquivalent2} ensures that this estimation exercise is not unboundedly misleading. Of note, since $\mu^b_t (a_t|\bPi_t^m)=\E^m_{\bS_t^m}\big[\mu^b_t (a_t|\bPi_t^m, \bS_t^m)\big]$, one can also view \eqref{eq:BUCEquivalent2} as bounded variations compared to the average (under each model).

Assuming that \eqref{eq:BUCEquivalent2} holds, we next extend Proposition \ref{prop:POMDPest} by relaxin the assumption that $\bmu^b$ satisfies sequential ignorability. To this end, we introduce the following {\em weight modifiers:}
\begin{equation}\label{eq:weightmod1}
\un \kappa^m_t\teq \Big(({\eta^m_t})^{-1}\, \1_{\{V_{T}^{\bmu^b} (\bpi) \geq 0\}}+ {\eta^m_t}\, \1_{\{V_{T}^{\bmu^b} (\bpi) <0\}}\Big)
\end{equation}
 and
\begin{equation}\label{eq:weightmod2}
\ov \kappa^m_t\teq \Big(({\eta^m_t})^{-1}\, \1_{\{V_{T}^{\bmu^b} (\bpi) <0\}}+ {\eta^m_t}\, \1_{\{V_{T}^{\bmu^b} (\bpi)\geq 0\}}\Big),
\end{equation}
and make use of the BUC results presented in Proposition \ref{prop:buc}.\footnote{These weight modifiers may depend on $\bpi$ in general (based on  \eqref{eq:weightmod1}-\eqref{eq:weightmod2}). Such dependency is suppressed here for the ease of notation.}

\begin{proposition}[Weight-Adjusted Bellman Equation Under BUC]\label{prop:POMDPestUBC}
Suppose $\bmu^b$ satisfies both the BUC condition \eqref{eq:BUCEquivalent2} and positivity. For any policy $\bmu^e$, define the upper and lower bound value functions via the modified weight-adjusted Bellman equations:
\begin{equation}\label{eq:POMDPest1BUC}
\ov V_{T-t+1}^{m,\bmu^e} (\bpi_t)\teq \E^m \bigg[\ov \kappa^m_t\,\frac{\mu_t^e(A_t|\bPi^m_t)}{\mu_t^b(A_t|\bPi^m_t)}\Big[G_t+ \beta\, \ov V_{T-t}^{m,\bmu^e}  (T(\bPi^m_t, A_t, O_t, m))\Big]\Big| \bPi^m_t=\bpi_t \bigg],
\end{equation}
and
\begin{equation}\label{eq:POMDPest2BUC}
\un V_{T-t+1}^{m,\bmu^e} (\bpi_t) \teq \E^m \bigg[\un \kappa^m_t\,\frac{\mu_t^e(A_t|\bPi^m_t)}{\mu_t^b(A_t|\bPi^m_t)}\Big[G_t+ \beta\, \un V_{T-t}^{m,\bmu^e}  (T(\bPi^m_t, A_t, O_t, m))\Big]\Big| \bPi^m_t=\bpi_t \bigg],
\end{equation}
along with
$\ov V_{0}^{m,\bmu^e} (\bpi)\teq 0$ and $\un V_{0}^{m,\bmu^e} (\bpi)\teq 0$. Then:
\begin{itemize}[leftmargin=1cm,align=left]
\item[(i)] For any function $\phi$ defined on $\Delta_\MS$, and for all $t\in\MT$ and $m\in \MM$, we have:
\begin{equation} \label{eq:POMDPest3BUC}
\E^m \bigg[\ov \kappa^m_t\, \frac{\mu_t^e(A_t|\bPi^m_t)}{\mu_t^b(A_t|\bPi^m_t)}\Big[G_t+ \beta\,  \ov V_{T-t}^{m,\bmu^e}  (T(\bPi^m_t, A_t, O_t, m))- \ov V_{T-t+1}^{m,\bmu^e} (\bPi^m_t)\Big] \phi (\bPi^m_t)\bigg]=0,
\end{equation}
and
\begin{equation} \label{eq:POMDPest4BUC}
\E^m \bigg[\un \kappa^m_t\, \frac{\mu_t^e(A_t|\bPi^m_t)}{\mu_t^b(A_t|\bPi^m_t)}\Big[G_t+ \beta\,  \un V_{T-t}^{m,\bmu^e}  (T(\bPi^m_t, A_t, O_t, m))- \un V_{T-t+1}^{m,\bmu^e} (\bPi^m_t)\Big] \phi (\bPi^m_t)\bigg]=0.
\end{equation}
\item[(ii)] For all $m\in \MM$ and $\bpi\in\Delta_\MS$ we have:  $ \un V_{T}^{m,\bmu^e} (\bpi) \leq V_{T}^{m,\bmu^e} (\bpi) \leq \ov V_{T}^{m,\bmu^e} (\bpi).$
\item[(iii)]  For any $\alpha\in\MI$, there exists $\tilde\alpha\in\MI$ such that
$V_{T}^{\bmu^e} (\bpi) \teq MEU_\alpha\big[V_{T}^{m,\bmu^e} (\bpi)\big]= f(\tilde\alpha, \bpi)$,
where $f (\tilde\alpha, \bpi)\teq \tilde\alpha\, MEU_\alpha\big[\un V_{T}^{m,\bmu^e} (\bpi)\big]+ (1-\tilde\alpha) MEU_\alpha\big[\ov V_{T}^{m,\bmu^e} (\bpi)\big].$ Hence, $\Gamma_T (\bmu^e)= \int f(\tilde\alpha, \bpi)\,dF(\bpi)$, where $\Gamma_T (\bmu^e)\teq \int V_{T}^{\bmu^e} (\bpi)\,dF(\bpi).$
\end{itemize}
\end{proposition}

When the behavior policy $\bmu^b$ satisfies sequential ignorability, we have $\un \kappa^m_t=\ov \kappa^m_t=1$, and hence, the above results boil down to those in Proposition \ref{prop:POMDPest}. Proposition \ref{prop:POMDPestUBC}, however, generalizes Proposition~\ref{prop:POMDPest} by highlighting the role of weight modifiers $\un \kappa^m_t$ and $\ov \kappa^m_t$  in analyzing scenarios in which $\bmu^b$ violates sequential ignorability, but satisfies it to {\em some extent}. In particular,  after incorporating these modifiers, one can make use of the same procedures as in the previous section using Proposition \ref{prop:POMDPestUBC}. That is, the sample-average version of \eqref{eq:POMDPest3BUC} and \eqref{eq:POMDPest4BUC} together with regularized learning can be used to learn the upper and lower bound value functions in a parametric way, respectively. Once these functions are learned, part (ii) of Proposition \ref{prop:POMDPestUBC} guarantees that they can be used to bound the actual value function under each model. More specifically, part (iii) of Proposition \ref{prop:POMDPestUBC} states that  $f(\tilde\alpha, \bpi)$ and $\int f(\tilde\alpha, \bpi)\,dF(\bpi)$ can be used to estimate the APOMDP value function and the overall performance, respectively. It should be noted that  (1) the function $f(\cdot, \cdot)$ is calculable using only observed part of the data. This resolves the issue that the outcome of interest under the evaluation policy as well as the time-varying confounders needed to estimate it are unobservable. (2) A similar procedure to that discussed after Proposition \ref{prop:buc} can be used to tune parameter $\tilde\alpha$. Specifically, as discussed there, noting that the parameters $(\eta_t^m)_{t\in\MT,m\in\MM}$ can be viewed as {\em design sensitivity parameters}, one can choose them and approximate $\tilde\alpha$ so as to obtain an approximate unbiased $MEU_\alpha$ estimator for  $\Gamma_T (\bmu^e)$ with any desired approximation error $\epsilon>0$.

Finally, the above results allow us to extend $\DAV$ and $\SAV$, and introduce their BUC counterparts. These extensions, which we term  $\DAVBUC$ and $\SAVBUC$, are presented in Algorithms  \ref{alg:DAVBUC} and \ref{alg:SAVBUC}, respectively (see Online Appendix C). The main difference between these extensions and their original version is that they benefit from weight modifiers \eqref{eq:weightmod1}-\eqref{eq:weightmod2} to first obtain estimators for the upper and lower bound value functions (as opposed to the main value function itself).

\section{Performance Analyses: Numerical Results}\label{sec:numerical}
To gain further insights into the performance of our purposed algorithms, we now perform two sets of numerical experiments. The first is a case study of a  medical decision-making problem faced by physicians at our partner hospital, and  involves using a clinical data set of
patients with a kidney transplant operation. In  the second set, we make use of synthetic data in which we simulate  patient trajectories under different models while controlling the true data generating model.

\subsection{Case Study: New Onset Diabetes After Transplantation (NODAT)}\label{sec:case}
In this section, we apply our proposed algorithms on a clinical data set that contains over  63,000 data points pertaining  407
patients who had a kidney transplant operation during a seven year period at our partner hospital.  Details about the data set can be found in the author's previous publications \citep{Boloori2015, Boloori2020, Munshi2020, Munshi2021}.

Patients who undergo transplantation often face a significant risk of organ rejection. To mitigate this risk, physicians typically use an intensive amount of an immunosuppressive drug (e.g., tacrolimus). Immunosuppressive drugs, however, have a well-established effect known as the diabetogenic effect,  and thus, can elevate the risk of {\em New Onset Diabetes After Transplantation (NODAT)}. NODAT refers to incidence of diabetes in a
patient with no history of diabetes prior to transplantation \citep[see, e.g., ][ and the references therein]{Chakkera2009, Boloori2015, Boloori2020}. To control the risk of NODAT, physicians have to decide whether or not to put the patient on insulin.\footnote{Of note,   similar to this study, \cite{Boloori2020} also address simultaneous management of immunosuppressive drugs and insulin for NODAT patients. However, the study of \cite{Boloori2020} is not concerned with the main aspects of this work. Namely, it does not deal with  (a) causal inference, or (b) Reinforcement Learning. The main ideas we use in this work are also not used in \cite{Boloori2020}. For example, we make use of {\em Importance Sampling} ideas along with  {\em weight-adjusted} versions of the Bellman equation,  but  in \cite{Boloori2020} the approach is vastly different: an APOMDP model is directly fitted to the data, and the optimal policy of this APOMDP is established using its Bellman equation (without weight adjustment) based on the theoretical results known for APOMDPs \citep{Saghafian2018}. We believe both the approach used in \cite{Boloori2020} and in this study are novel. However, they are not directly comparable in a ceteris paribus manner due to the above-mentioned differences.}

Table \ref{Table:Observations} describes the observed patient covariates (observations) and their levels. As the table shows, some of these observations are time-varying. Furthermore, most of them are dichotomized to high versus low level values. However, the medical tests  used to measure the blood glucose (FPG and Hb1Ac) and the lowest concentration of tacrolimus
in the patient's body---a quantity known as {\em trough level} or $C_0$---have three levels. These levels are defined based on both the medical literature and the practice at our partner hospital. Tables \ref{Table:States}  and \ref{Table:Actions} show the patients' latent states and physicians' actions/prescriptions during each visit post-transplant, respectively.   Latent states described in Table \ref{Table:States} are summary variables that describe the main condition of the patient in terms of decision-making related to use of an immunosuppressive drug (e.g., tacrolimus) and insulin therapy (i.e., the actions in  Table~\ref{Table:Actions}). These patient summary variables are, however, hidden to physicians,  since physicians can only rely on medical tests, which have a wide range of false-positive and false-negative errors. In particular, blood glucose levels are measured by two medical tests {\em Fasting Plasma Glucose} (FPG) and {\em Hemoglobin A1c} (HbA1c), which are subject to false-positive and false-negative errors. Similarly, the concentration of immunosuppressive drugs is measured through tests such as {\em Abbott Architect} and {\em Magnetic Immunoassay}, which are error-prone.

\noindent\textbf{Data Pre-processing Steps.} Our data set  includes information related to patients' follow-up visits during  months 1, 4, and 12 post transplantation. However, for the goals of this study,  we make use of the same data preprocessing steps as those in \citep{Boloori2020}. In particular, we  use imputation to replace missing values \citep[see also][]{Munshi2021} and also make use of cubic spline interpolation to create a test bed with clinical history of patients for months 1 to
12 after transplant. That is, for the purpose of this study, we consider monthly visits that occur for a year post-transplant. Thus, we let $T\teq 12$ and $\MT\teq\{1,2,\cdots, 12\}$. The imputed data includes the 13 variables listed in  Table \ref{Table:Observations} for each of the 407 patients and every month during a year of follow-up post-transplant (a total of $13\times 407\times 12=63,492$ data points).

{
	\renewcommand{\arraystretch}{1.2}
	\begin{table}[t]
		\caption{\baselineskip=10pt  Observed Covariates (Observations)}\label{Table:Observations}

		\begin{center}
			\scriptsize
			\resizebox{\textwidth}{!}{\begin{tabular}{clccccc}
				\hline
			{\bf Var. No.} &	{\bf Risk Factor (Abbr.)} & {\bf Unit} & {\bf Low Level} & {\bf Mid Level} & {\bf High Level} & {\bf Time-Varying} \\ \hline
			1                  & 	Glucose test$^\dag$ (FPG, HbA1c) &  mg/dL, \% & Healthy &  Pre-Diabetic & Diabetic&  Yes \\
            2                  &    Trough level test$^\ddag$ ($C_0$)  & mg/dL &$[4,8)$ &$[8,10)$& $[10,14]$& Yes \\
            3                  &    Age & Years & $<$50 & --- & $\geq$ 50 & No \\
			4                  &	Gender & --- & Female &---  & Male & No \\
			5                  &	Race & --- & White & --- & non-White & No \\
			6                  &	Diabetes history (Diab Hist) & --- & No & --- & Yes & No \\
			7                  &	Body mass index (BMI) & kg/m$^2$ & $<$30 (non-obese) & --- & $\geq$30 (obese) &  Yes \\
			8                  &	Blood pressure (BP) & --- & Normal$^\sharp$ &--- &  Hypertension &  Yes \\
			9                  &	Total cholesterol (Chol) & mg/dL & $<$200 & --- & $\geq$200 &  Yes \\
			10                 &	High-density lipoportein (HDL) & mg/dL &  $\geq$40 & --- & $<$40 &  Yes\\
			11                 &	Low-density lipoportein (LDL) & mg/dL &  $<$130 & --- & $\geq$130 &  Yes \\
			12                 &	Triglyceride (TG) & mg/dL &  $<$150 & --- & $\geq$150 &  Yes \\
			13                 &	Uric acid (UA) & mg/dL &  $<$7.3 & --- & $\geq$7.3 &  Yes \\
				
\hline
\multicolumn{6}{l}{$^\dag$\normalfont A patient with FPG$\geq$126 ($100\leq$FPG$<126$) mg/dL or HbA1c$\geq$6.5\% ($5.7\leq$HbA1c$<$6.5\%) is labeled as diabetic (pre-diabetic),}\\
 \multicolumn{6}{l}{and a patient with FPG$<$100 mg/dL or HbA1c$<$5.7\% is labeled as healthy \citep[see, e.g.,][]{ada2012}.}\\
\multicolumn{6}{l}{$^\ddag$\normalfont $C_0 \in [4,8)$, $[8,10)$, $[10,14]$ mg/dL is label as ``low,'' ``medium,'' and ``high,'' respectively \citep[see, e.g.,][]{Boloori2020}.}\\
\multicolumn{6}{l}{$^\sharp$\normalfont Normal Blood Pressure (BP) is defined as systolic (diastolic) BP less than 120 (80) mmHg \citep[see, e.g.,][]{Whelton2018}.}\\
\multicolumn{6}{l}{\normalfont Note: All variables with three levels are coded as 1,2, 3 (low, mid, high). All variables with two levels are coded as 1, 2 (low,  high).}\\
			\end{tabular}}

		\end{center}
		\vspace{-10pt}
	\end{table}
}

\noindent\textbf{Behavior Policy.} We estimate the behavior policy based on the actions we observe in our data. These actions are mainly based on the the clinical protocols followed at our partner hospital. A detailed summary of the main immunosuppression protocol can be found in \citep{Munshi2021}, which includes induction therapy with
either rabbit anti-thymocyte, immunoglobulin, or basiliximab, as well as a tapering course of glucocorticoids. However, here our focus  is on the use of tacrulimus, and we observe that  patients are often put on high (i.e., aggressive) dose tacrolimus during the first months post-transplant, and in later months, depending on the observations made about the patient patients, they might be transferred to a  low (i.e., non-aggressive) dose.  This is consistent with the fact that patients in most medical practices are consistently kept on high levels of tacrolimus
in early stages post-transplant  \citep[see, e.g., ][]{Ghisdal2012, Boloori2020}. Furthermore, with respect to the use of insulin,  patients are primarily put on insulin when their Hb1Ac and FPG tests indicates that they are not diabetic free (see definitions of pre-diabetic and diabetic in Table \ref{Table:Observations}). Using the observed actions in our data set as well as the estimated belief vectors $\{\bpi_t^m\}_{t\in\MT}$ for each patient (for further details, see the ``Other Details" paragraph below),  we next estimate $\mu^b(A_t|\bPi^m_t)$ by training  a multi-class  multiple logistic regression classifier. This classifier  is endowed with an $\ell_2$-norm penalty, which is tuned to ensure that each action is selected with an estimated probability of $0.05$ or higher across all observations \citep[see, e.g., ][]{Murphy2016}.

{
	\renewcommand{\arraystretch}{1.2}
	\begin{table}[t]
				\caption{\baselineskip=10pt Latent Health States}\label{Table:States}
				\begin{center}
\scriptsize
		   \begin{tabular}{ccc}
				\hline
				\multirow{2}{*}{{\bf State}} & \multicolumn{1}{c}{\bf Transplant Condition} & \multirow{2}{*}{{\bf Diabetes Condition}} \\
				& \multicolumn{1}{c}{\bf (Tacrolimus $C_0$)} & \\
				\hline
				$1$ & Low & \multirow{3}{*}{Diabetes (type II)} \\
				$2$ & Medium &  \\
				$3$ & High &  \\ \hline
				$4$ & Low & \multirow{3}{*}{Pre-diabetes} \\
				$5$ & Medium & \\
				$6$ & High & \\ \hline
				$7$ & Low & \multirow{3}{*}{Healthy} \\
				$8$ & Medium & \\
				$9$ & High & \\
				\hline
				
			\end{tabular}
		\end{center}
	\end{table}
}

{
\renewcommand{\arraystretch}{1.2}
	\begin{table}[b]
				\caption{\baselineskip=10pt Actions}\label{Table:Actions}
		\begin{center}
			\scriptsize
			\begin{tabular}{ccc}
				\hline
				\multirow{2}{*}{{\bf Action}} & \multicolumn{1}{c}{\bf Prescription} & \multicolumn{1}{c}{\bf Prescription} \\
				& \multicolumn{1}{c}{\bf (Tacrolimus dose)} & \multicolumn{1}{c}{\bf (Insulin use)} \\ \hline
				$1$ & Low (Non-Aggressive) & \multirow{2}{*}{No} \\
				$2$ & High (Aggressive) & \\ \hline
				$3$ & Low (Non-Aggressive) & \multirow{2}{*}{Yes} \\
				$4$ &High (Aggressive)& \\
				\hline
				& & \\ \\
				& & \\ \\
			\end{tabular}
		\end{center}
	\end{table}
}

\noindent\textbf{Immediate Gain Variable.} To calculate the immediate gains, we use a similar approach to our previous work \citep[see, e.g.,][]{Boloori2020}. In particular, we make use of {\em Quality of Life (QoL) scores}, which take values in $[0,1]$. This allows us to differentiate between the Quality of Life of being in a diabetic, prediabetic,  or healthy state and also having different concentration of the immunosuppressive in the body, which are in turn associated with differing risks of organ rejection. Table  \ref{Table:Gain} shows the yearly-based QoL scores associated with each state, which are divided by 12 to represent the fact that patients' visits are monthly.\footnote{In addition to immediate gains, our framework  allows including lump-sum gains (i.e., gains at the end of the horizon to reflect the Quality of Life associated with the remaining years). For the purposes of this study, however, we simply set $V_0(\bpi)\teq 0$.}\footnote{Given $\bpi^m_t$ in each period $t$ under each model $m$, the obtained immediate gain in each period $t$ under each model $m$ is considered be the weighted average of immediate gain values shown in Table  \ref{Table:Gain}, where weights are given by $\bpi^m_t$ (see also Remark \ref{remark:BUC}).}

{
	\renewcommand{\arraystretch}{1.2}
	\begin{table}[t]
				\caption{\baselineskip=10pt Immediate Gain Values}\label{Table:Gain}
				\begin{center}
\scriptsize
		   \begin{tabular}{cccc}
				\hline
				\multirow{2}{*}{{\bf State}} & \multicolumn{1}{c}{\bf Transplant Condition} & \multirow{2}{*}{{\bf Diabetes Condition}}& \multirow{2}{*}{{\bf Immediate Gain Value$^\dag$}} \\
				& \multicolumn{1}{c}{\bf (Tacrolimus $C_0$)} & \\
				\hline
				$1$ & Low & \multirow{3}{*}{Diabetes (type II)} & $0.68/12$\\
				$2$ & Medium &  &$0.72/12$\\
				$3$ & High &  & $0.76/12$\\ \hline
				$4$ & Low & \multirow{3}{*}{Pre-diabetes}& $0.82/12$ \\
				$5$ & Medium & &$0.87/12$\\
				$6$ & High & & $0.89/12$\\ \hline
				$7$ & Low & \multirow{3}{*}{Healthy} &$0.90/12$\\
				$8$ & Medium & &$0.92/12$\\
				$9$ & High & &$0.95/12$\\
				\hline
\multicolumn{4}{l}{$^\dag$\normalfont Immediate gains are average values approximated based on $QoL$ scores reported in other studies and include combined}\\
\multicolumn{4}{l}{disutility of (a) being in a diabetic state, and (b) having high risk of organ rejection. Yearly-based values are divided by}\\
		\multicolumn{4}{l}{12 to represent monthly measures.}
			\end{tabular}
		\end{center}
	\end{table}
}

\noindent\textbf{Other Details.} The belief state space in our setting,   $\Delta_\MS$, is a $8$-simplex, since there are 9 latent states (Table \ref{Table:States}). The vector of basis functions $\bb(\bpi)$ maps this  $8$-simplex to $\mathbb R^{13}$, which allows us to include enough cut points (while making sure that the value function is piecewise linear and continuous).  Thus, both the belief space and the parameter space in our setting are continuous and relatively high-dimensional. To perform our analyses, we use a discount factor of $\beta=0.95$.  We also tune a  penalty parameter $\theta_t=\theta$. To create the set of models $\MM$, we  make use of  the algorithm in Table 3 of our earlier work \citep{Boloori2020}. Specifically, first the Baum–Welch algorithm is used to obtain point estimations for  state transition and observation probability matrices. Next, an entropy ball is constructed (using the Kullback–Leibler divergence criterion) around these point estimate matrices. For tractability, we set $|\MM|=4$ in this case study. However, our framework is general and can be used for any number of estimated models.  In Section \ref{sec:synthetic}, for example, we change our assumption on the number of models  and consider $|\MM|=10$ different models. Our framework is also not restricted to any specific way of estimating the underlying models. For example, in Section \ref{sec:synthetic}, we make use of a  different way of constructing the set $\MM$.\footnote{It should be also noted that any continuous set of models can be approximated via finite sets
with any required precision. That is, even if $\MM$ is not finite, one can always consider a finite set $\MM$ as a close approximation to the continuous one.} Finally, we consider the distribution $F(\bpi)$ to be uniform.  That is, we use a uniform prior belief at time zero, and implement  the Bayesian belief updating operator  (see Eq. \eqref{beliefupd})  to create a sequence of belief vectors $\{\bpi_t^m\}_t\in\MT$ for each patient under each model $m\in\MM$ (see, e.g., steps 1-5 in Algorithms \ref{alg:DAV}, \ref{alg:SAV}, \ref{alg:DAVBUC}, and \ref{alg:SAVBUC}).

\noindent\textbf{Results.} The performance of the three treatment regimes ($\DAV$, $\SAV$, and observed) are compared in Table \ref{Table:Results1}.  Average and standard deviations in these tables are calculated using Monte Carlo replications.\footnote{The number of these replications is chosen so that the confidence intervals are tight enough, while maintaining  reasonable computational times.}  We focus on the performance of $\DAV$, $\SAV$ as opposed to their BUC extensions  ($\DAVBUC$, $\SAVBUC$), because as discussed in Remark \ref{remark:BUC}, the results of Proposition \ref{prop:POMDPest} hold in this application. In the next section, we run experiments using $\DAV$ and $\SAV$ as well as $\DAVBUC$ and $\SAVBUC$.

\begin{figure}[t]
  \begin{center}
  \includegraphics[scale=0.5]{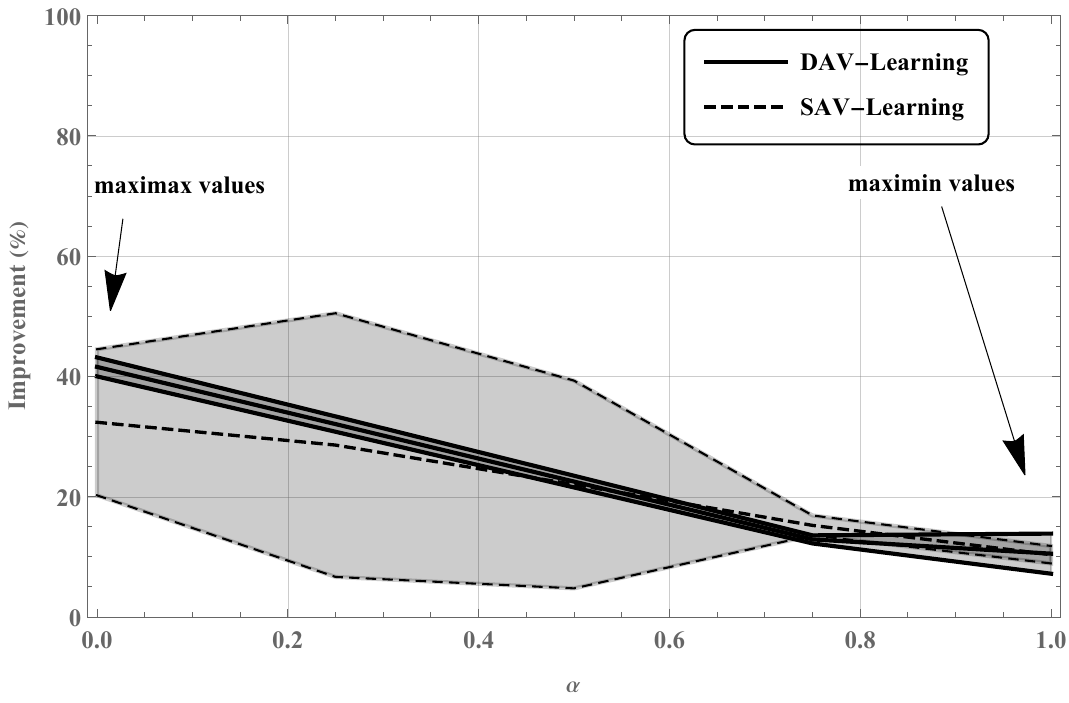}\vspace{2mm}
  \caption{\scriptsize Percentage improvement over  the observed regime (case study with $\beta=0.95$). Gray areas represent error bands with the curve at the center of each error band representing the mean value.}\label{fig:improvementcase}
    \end{center}\vspace{-4mm}
\end{figure}

As can be seen from the results in  Table \ref{Table:Results1}, $\DAV$  outperforms $\SAV$ in terms of the mean performance for most values of the pessimism level, $\alpha$. As both  Table \ref{Table:Results1} and Figure~\ref{fig:improvementcase}  show, however, both $\DAV$ and $\SAV$ approaches significantly outperform the observed regime. In particular, as Figure \ref{fig:improvementcase} shows, the improvements over the observed regime when using $\DAV$ and $\SAV$ are in the ranges $(10\%, 42\%)$ and $(10\%, 32\%)$, respectively, depending the value of $\alpha$. Of note, these ranges also imply  that the mean performance of the $\SAV$ regime is much more robust to the value of $\alpha$ than that of $\DAV$. This is due to the fact that $\SAV$ uses a ``safe estimation" of the  underlying parameter of the value function (see, e.g., step 12 of Algorithm \ref{alg:SAV}). This allows $\SAV$ to guard against ambiguity up-front (i.e., in parameter estimation) in contrast to $\DAV$ which combines  policy values at the end. Thus, a decisions-maker who uses $\SAV$ does not need to be that concerned about the value of $\alpha$ s/he uses (or try to tune it).

Finally, as can be seen from both  Table \ref{Table:Results1} and Figure \ref{fig:improvementcase}, the performance of $\DAV$  and $\SAV$ regimes degrades as the pessimism level $\alpha$  increases. This is fully expected, since as we move from a maximax view to a maximin one  $\DAV$  and $\SAV$  tend to put more weight on the worst-case scenario, and hence, perform more {\em conservatively}. More conservativeness, however, does not necessarily mean more {\em robustness} to model ambiguity. We further investigate this issue in Section
\ref{sec:robsutness}, and generate important insights into the values of $\alpha$ that can provide the highest level of robustness to model ambiguity.

{
	\renewcommand{\arraystretch}{1.2}
	\begin{table}[t]
		\caption{\baselineskip=10pt  Estimated Total Discounted Gain Under Observed and Proposed Regimes (Case Study with $\beta=0.95$)}\label{Table:Results1}

		\begin{center}
			\scriptsize
			\begin{tabular}{cccc}
				\hline
				{\bf Pessimism Level ($\alpha$)} & {\bf Observed Regime$^\dag$} & {\bf $\DAV$$^\dag$} & {\bf $\SAV$$^\dag$} \\ \hline
				0.00 &  1.472 (1.455, 1.489)  & {\bf 2.085} (2.061, 2.108)  & 1.949 (1.770, 2.128) \\
                0.25 &  1.468 (1.456, 1.480)  & {\bf 1.939} (1.920, 1.958)  & 1.888 (1.566, 2.210) \\
                0.50 &  1.464 (1.457, 1.471)  & {\bf 1.794} (1.779, 1.808)  & 1.786 (1.534, 2.039) \\
                0.75 &  1.460 (1.458, 1.462)  & 1.648 (1.638, 1.658)  & {\bf 1.682} (1.658, 1.706) \\
				1.00 &  1.455 (1.452, 1.458)  & {\bf 1.609} (1.560, 1.657)  & 1.606 (1.585, 1.627) \\
\hline
\multicolumn{4}{l}{$^\dag$\normalfont Values in parenthesis represent $95\%$ confidence intervals. Values in bold font represent the best performance.}\\
\multicolumn{4}{l}{\normalfont For all values, only the first three decimal places are shown.}\\
			\end{tabular}
		
		\end{center}
		\vspace{-10pt}
	\end{table}
}

\subsection{Synthetic Data Analyses}\label{sec:synthetic}
We now use similar assumptions to those described in the case study, but instead of using actual patient traceries, simulate random patient trajectories for 100 patients with 10 follow-up periods, and use ($|\MM|=10$) different models. These yield randomly generated belief data of the form $(\bpi^m_t)_{t\in\MT}$ under each $m \in\MM$.  We keep the other assumptions (e.g., the action space, the number of hidden states,  the parameter space, basis functions, etc.) the same as those in the previous section.

We assume patient trajectories are such that for each $m\in\MM$ the belief vector $(\bpi^m_t)_{t\in\MT}$ is generated via a Dirichlet distribution with the vector of parameters $(p^m_i)_{i\in\{1,2\cdots,9\}}$. All of these models are misspecified, and hence,  for each model, we randomly draw each $p^m_i$ from a Uniform$(0,1)$ distribution.  We assume the true model is such that all $p_i$ values are equal to 0.5. Furthermore, we specify the behavior policy as follows. For actions $a=1,2,3$, we set $\mu^b(A=a|\bPi=\bpi)=\frac{exp (\bpi' \, \bvarrho_a)}{1+\sum_{a=1}^3 exp (\bpi' \, \bvarrho_a)}$ and for action $a=4$ we set $\mu^b(A=a|\bPi=\bpi)=\frac{1}{1+\sum_{a=1}^3 exp (\bpi' \, \bvarrho_a)},$ where  $\bvarrho_1$, $\bvarrho_2$, and $\bvarrho_3$ are 9-dimensional predefined vectors. To perform our analyses, we choose  each $\bvarrho_a$ ($a=1,2,3$) as a vector with all elements equal to $0.1$, except the $a$-th element, which is set to $-1$.

\begin{figure}[h]
  \begin{center}
  \includegraphics[scale=0.5]{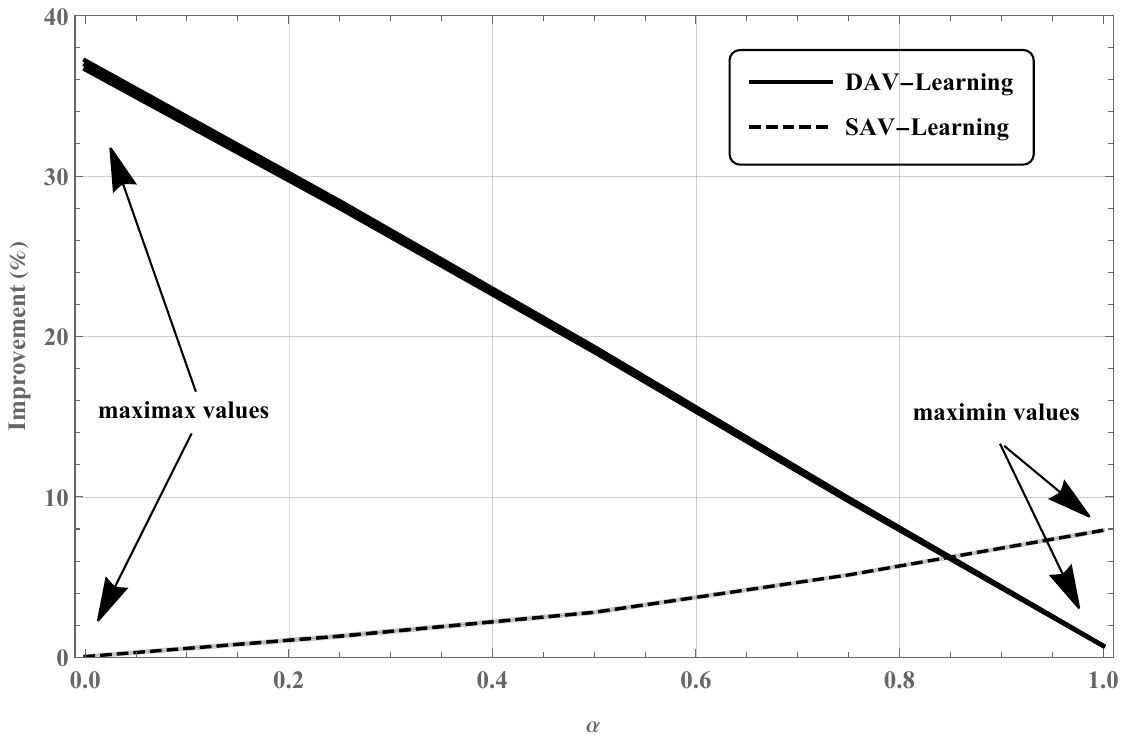}\vspace{2mm}
  \caption{\scriptsize Percentage improvement over  the observed regime (synthetic data analyses with $\beta=0.95$). Error bands for both approaches, and especially for the $\SAV$ approach, are very tight (hence, not depicted).}\label{fig:improvementsynthetic}
    \end{center}\vspace{-4mm}
\end{figure}

\begin{figure}[h]
  \begin{center}
  \includegraphics[scale=0.5]{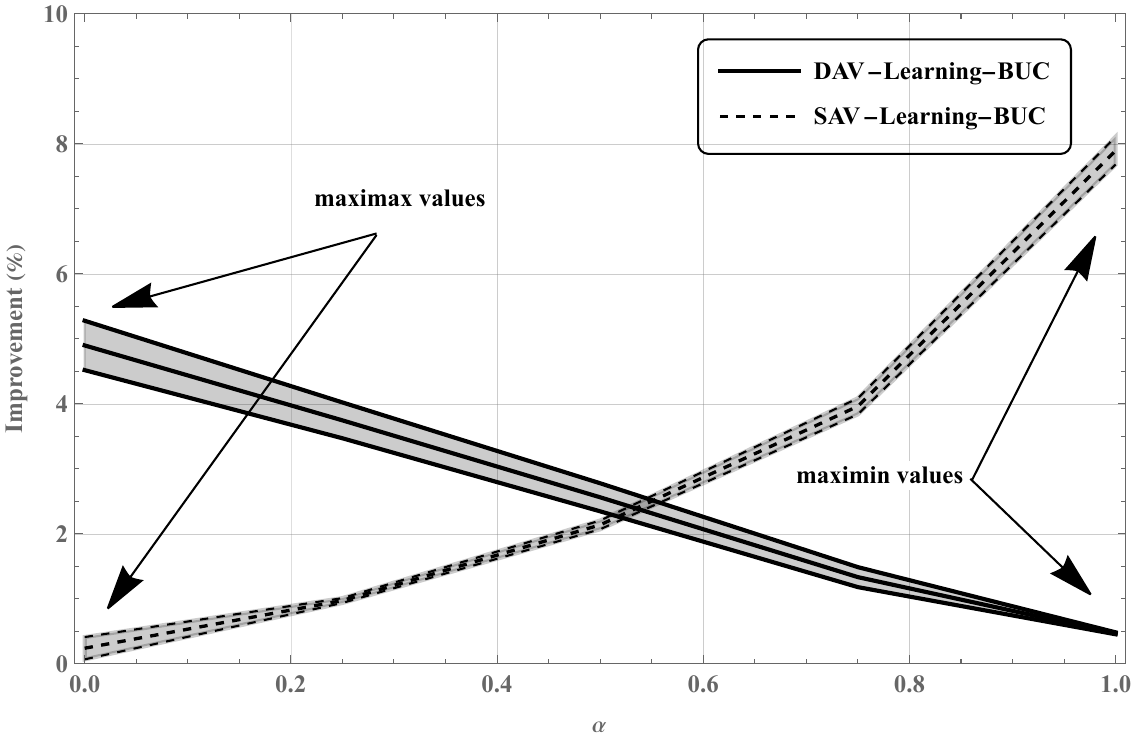}\vspace{2mm}
  \caption{\scriptsize Percentage improvement over  the observed regime (synthetic data analyses with $\beta=0.95$ and $\eta_t^m=1.02$ for all $t\in\MT$ and $m\in\MM$).}\label{fig:improvementsyntheticBUC}
    \end{center}\vspace{-4mm}
\end{figure}

Table \ref{Table:Results2} and Figure \ref{fig:improvementsynthetic} (Online Appendix D) present our results using the $\DAV$ and $\SAV$ approaches under the same immediate gain values as those in the case study (see Table  \ref{Table:Gain}). Similar results for the BUC version of these approaches, $\DAVBUC$ and $\SAVBUC$, are presented in Table \ref{Table:Results2BUC} and Figure \ref{fig:improvementsyntheticBUC} (Online Appendix D). Similar to the case study, we observe from these results that  all of our  proposed algorithms outperform the observed regime. Furthermore, the percentage improvement of $\DAV$ an $\SAV$ over the observe regime ranges in $(1\%, 37\%)$  and $(1\%, 8\%)$, respectively, depending on the value of $\alpha$.  These improvement ranges for $\DAVBUC$ an $\SAVBUC$ are $(0.7\%, 4.9\%)$ and $(0.07\%, 7.9\%)$, respectively. In addition, similar to our observation in the case study, $\DAV$  outperforms $\SAV$  for most vales of $\alpha$ (Figure \ref{fig:improvementsynthetic}), and a similar result can be seen for the BUC regimes (Figure \ref{fig:improvementsyntheticBUC}). Furthermore, similar to the case study, the performance of $\SAV$   is much more robust to the value of $\alpha$ compared to $\DAV$, but in the BUC regimes, $\DAVBUC$ shows relatively more robustness than $\SAVBUC$. In the next section, we further investigate the robustness of our proposed approaches to model ambiguity, and generate insights into the best value of   $\alpha$ that a decision-maker can use to achieve the highest level of robustness.

{
	\renewcommand{\arraystretch}{1.2}
	\begin{table}[t]
		\caption{\baselineskip=10pt  Estimated Total Discounted Gain Under Observed and Proposed Regimes  (Synthetic Data Analyses with $\beta=0.95$)}\label{Table:Results2}

		\begin{center}
			\scriptsize
			\begin{tabular}{cccc}
				\hline
				{\bf Pessimism Level ($\alpha$)} & {\bf Observed Regime$^\dag$} & {\bf $\DAV$$^\dag$} & {\bf $\SAV$$^\dag$} \\ \hline
				0.00 &  1.441 (1.440, 1.442)  & {\bf 1.973} (1.969, 1.977)  & 1.442 (1.441, 1.442) \\
                0.25 &  1.415 (1.415, 1.416)  & {\bf 1.815} (1.811, 1.818)  & 1.434 (1.433, 1.434) \\
                0.50 &  1.389 (1.389, 1.390)  & {\bf 1.656} (1.654, 1.659)  & 1.428 (1.428, 1.429) \\
                0.75 &  1.364 (1.364, 1.364)  & {\bf 1.498} (1.496, 1.499)  & 1.434 (1.434, 1.434) \\
				1.00 &  1.338 (1.338, 1.339)  & 1.348 (1.348, 1.348)  & {\bf 1.444} (1.444, 1.444) \\
\hline
\multicolumn{4}{l}{$^\dag$\normalfont Values in parenthesis represent $95\%$ confidence intervals. Values in bold font represent the best performance.}\\
\multicolumn{4}{l}{\normalfont For all values, only the first three decimal places are shown.}
			\end{tabular}
		
		\end{center}
		\vspace{-10pt}
	\end{table}
}

{
	\renewcommand{\arraystretch}{1.2}
	\begin{table}[t]
		\caption{\baselineskip=10pt  Estimated Total Discounted Gain Under Observed and Proposed BUC Regimes  (Synthetic Data Analyses with $\beta=0.95$)}\label{Table:Results2BUC}

		\begin{center}
			\scriptsize
			\begin{tabular}{cccc}
				\hline
				{\bf Pessimism Level ($\alpha$)} & {\bf Observed Regime$^\dag$} & {\bf $\DAVBUC$$^\dag$$^\ddag$} & {\bf $\SAVBUC$$^\dag$$^\ddag$} \\ \hline
				0.00 &  1.441 (1.440, 1.442)  & {\bf 1.511} (1.506, 1.517)  & { 1.448} (1.442, 1.447) \\
                0.25 &  1.415 (1.415, 1.416)  & {\bf 1.468} (1.464, 1.472)  & { 1.429} (1.428, 1.429) \\
                0.50 &  1.389 (1.389, 1.390)  & {\bf 1.425} (1.422, 1.428)  & { 1.419} (1.418, 1.420) \\
                0.75 &  1.364 (1.364, 1.364)  &  1.382 (1.380, 1.384)       & {\bf 1.418} (1.416, 1.420) \\
				1.00 &  1.338 (1.338, 1.339)  &  1.344 (1.344, 1.345)       & {\bf 1.444} (1.441, 1.447) \\
\hline
\multicolumn{4}{l}{$^\dag$\normalfont Values in parenthesis represent $95\%$ confidence intervals. Values in bold font represent the best performance.}\\
\multicolumn{4}{l}{\normalfont For all values, only the first three decimal places are shown.}\\
\multicolumn{4}{l}{$^\ddag$\normalfont Algorithm is run by assuming $\eta_t^m=1.02$ for all $t\in\MT$ and $m\in\MM$. Results are based on $\epsilon$-approximations}\\
 \multicolumn{4}{l}{for a small $\epsilon$ (see the discussion in Section \ref{sec:extension}). }
			\end{tabular}
		
		\end{center}
		\vspace{-10pt}
	\end{table}
}

\subsection{Robustness to Model Ambiguity}\label{sec:robsutness}
We now compare our proposed approaches in terms of their percentage {\em gain loss} (a.k.a., {\em regret}). That is, we first consider an {\em oracle} who knows both the true data generating model and the optimal policy under it, and then compare the performance of a decision-maker who is blind to the true data generating model (is facing model ambiguity) but uses either  $\DAV$ or $\SAV$ (or their BUC version, $\DAV$ or $\SAV$).  How much robustness to model ambiguity using  these proposed approaches provide? What is the maximum gain loss of these approaches? For what value of $\alpha$ the gain loss is minimized? Importantly, in order to minimize the gain loss, should the decision-maker use an extreme value of $\alpha$ (e.g., $\alpha=0,1$) or a mid level value (e.g., $\alpha=0.5$)? And does the answer depend on which learning approach is used?

\begin{figure}[h]
  \begin{center}
  \includegraphics[scale=0.5]{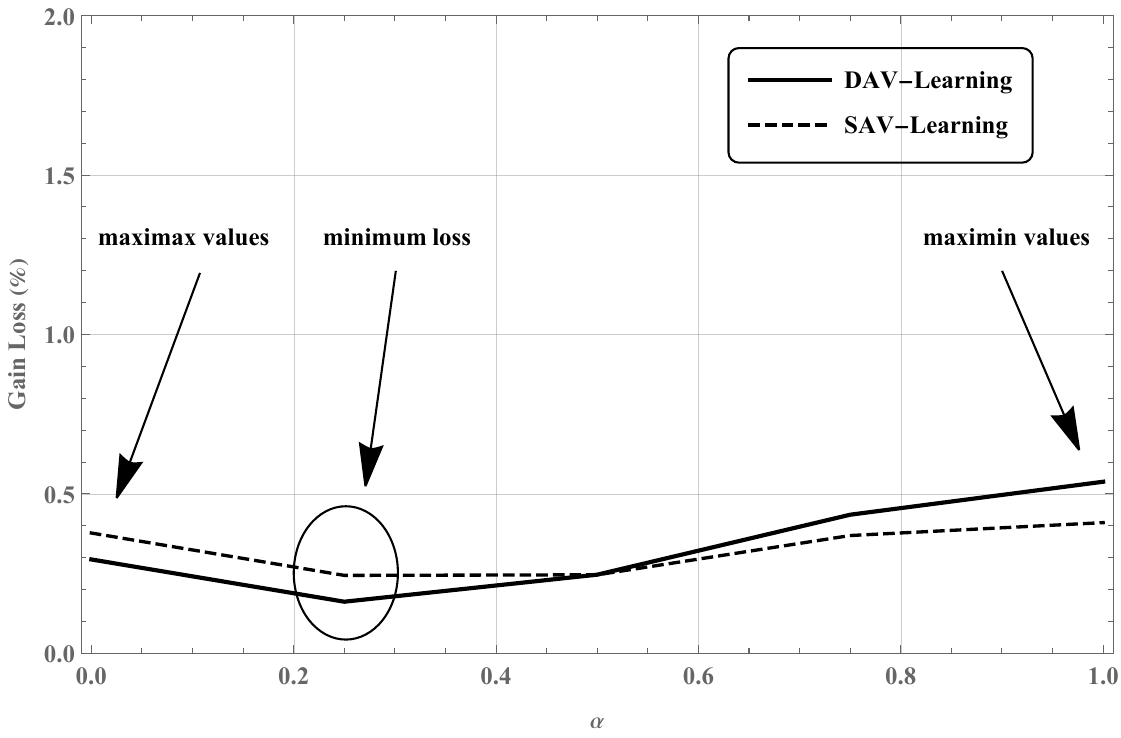}\vspace{2mm}
  \caption{\scriptsize Percentage Gain Loss of $\DAV$ and $\SAV$  (Synthetic Data Analyses with $\beta=0.95$). Minimum loss  is obtained for a mid level value of the pessimism level ($\alpha=0.25$).}\label{fig:robustness}
    \end{center}\vspace{-4mm}
\end{figure}

\begin{figure}[h]
  \begin{center}
  \includegraphics[scale=0.5]{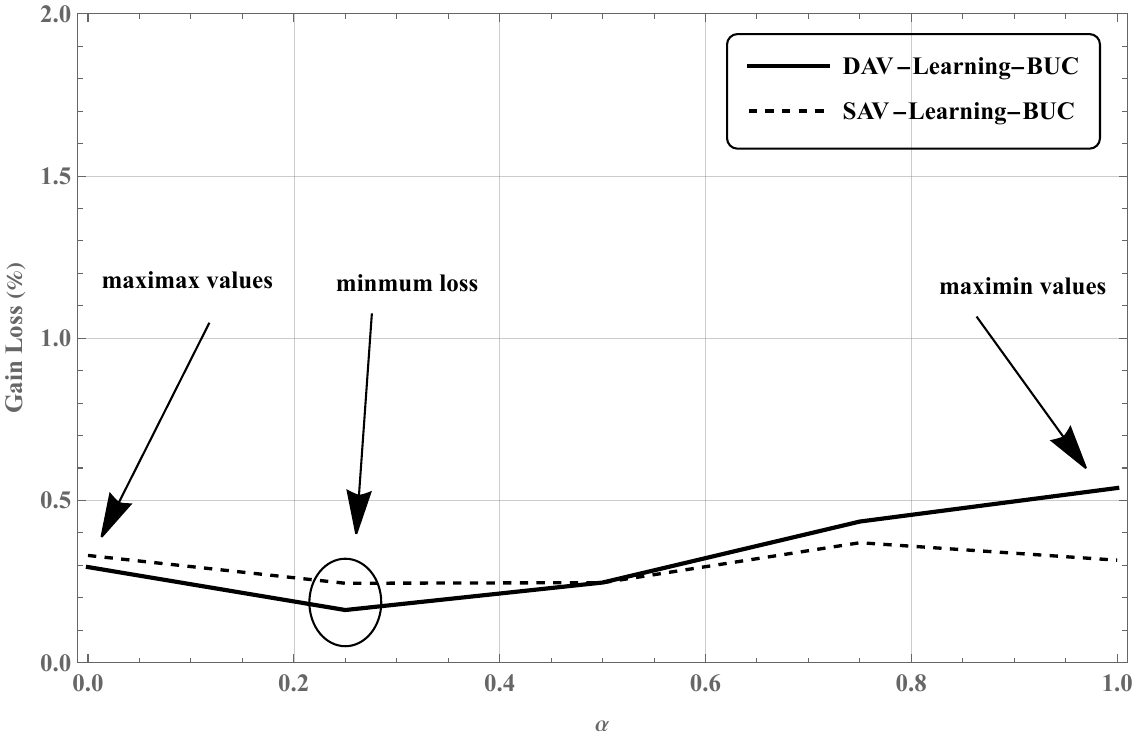}\vspace{2mm}
  \caption{\scriptsize Percentage Gain Loss of $\DAVBUC$ and $\SAVBUC$  (Synthetic Data Analyses with $\beta=0.95$ and $\eta_t^m=1.02$ for all $t\in\MT$ and $m\in\MM$). Minimum loss  is obtained for a mid level value of the pessimism level ($\alpha=0.25$). Results are based on $\epsilon$-approximations for a small $\epsilon$ (see the discussion in Section \ref{sec:extension}))}\label{fig:robustnessBUC}
    \end{center}\vspace{-4mm}
\end{figure}

To answer these  questions, we make use of a similar setup to the one discussed in Section~\ref{sec:synthetic}. The results are shown in Figure \ref{fig:robustness} (Online Appendix D), which depicts the percentage gain loss of  $\DAV$ and  $\SAV$  compared to the imaginary oracle. Similar results for the BUC version of these approaches ($\DAVBUC$ or $\SAVBUC$) are provided in  in Figure \ref{fig:robustnessBUC} (Online Appendix D). From these figure, we make three main observations: (1) Gain loss  has a U-shape curve as $\alpha$ varies. Importantly, the minimum loss for all four approaches ($\DAV$, $\SAV$ $\DAV$, $\SAV$)  are obtained at a mid value of $\alpha$ (approximately $\alpha=0.25$), which implies that using extreme cases of $\alpha=0.0$ (a maximax view) or $\alpha=1.0$ (a maximin view) does not provide the highest level of robustness to model ambiguity. That is, neither the maximax view nor the maximin view is {\em robustness-maximizing}. (2) The gain loss under $\SAV$ ($\SAVBUC$) is more robust to the changes in value of $\alpha$ compared to $\DAV$ ($\DAVBUC$). (3) All four proposed approaches are able to strongly shield against model ambiguity, regardless of the value of  $\alpha$ used. Specifically, the gain loss under these approaches (compared to the imaginary oracle) is very low (below $0.6\%$). This implies that a decision-maker who is facing model ambiguity can use these approaches and obtain  policies that have similar performance to the very best policy that could be used, if  the true data generating was known (i.e., if there was no ambiguity regarding the underlying causal model).\footnote{These findings hold in the context of our numerical experiments. However, we avoid making general conclusions, since doing so will require a more extensive set of experiments. Our results, however, provide a proof of concept that can be further explored by future research.}

\begin{remark}[\textbf{Dimensionality and Computations}]
The proposed algorithms $\DAV$ and $\SAV$ as well as their BUC extensions,   $\DAVBUC$ and $\SAVBUC$,  do not require their main parameter spaces to have low dimensionality. That is,  they can safely be used in high-dimensions, especially because they make use of regularization to avoid overfitting in high-dimensional settings.  In particular, both our theoretical and numerical performance results (Sections \ref{sec:assymptotics} and \ref{sec:numerical}) indicate that these algorithms have suitable convergence results and are relatively tractable for use in real-world applications. For example,  while the results in Section \ref{sec:assymptotics} indicate that only some typical ``regularity" conditions are needed to ensure that they have suitable asymptotic convergence behavior, numerical experiments in Section \ref{sec:numerical}  suggest that they are indeed tractable in real-world applications. However, it should be noted that our numerical experiments are motivated by a specific medical decision-making application in which the belief state space,   $\Delta_\MS$, is a $8$-simplex (since there are 9 latent states), the vector of basis functions $\bb(\bpi)$ maps this  $8$-simplex to $\mathbb R^{13}$, and the main parameter needed to estimate the value function has a reasonable dimensionality (belongs to $\mathbb R^{13}$). Furthermore, the set of models and the  policies needed in this application are not extremely large. If, in an application, the cardinality of these underlying spaces is significantly higher than those in our experiments, the proposed algorithms may lose computational tractability. In such scenarios, further care (e.g., discretization, approximation, etc.) is needed to speed up these algorithms. Finally,  it should be noted that the proposed algorithms are suitable for scenarios in which the observed history provides at least some information about latent confounders. In APOMDPs, some models might be naturally more informative than others; see Definition 2 and Lemma 2 of \cite{Saghafian2018} for the notion of model informativeness in APOMDPs, which is based on the Blackwell–Sherman–Stein sufficiency theorem. However, if none of the models are informative, then the Bayesian operator that is used in these algorithms might fail to update the belief distributions under all models, and hence, the value function cannot be learned from data for different values of $\bpi$. Intuitively, if the observed data does not provide any information about the dynamic latent confounders, then one should not hope for using observed variables to effectively adjust for the effect of dynamic unobserved confounders.
\end{remark}

\section{Conclusion}\label{sec:conclude}
We propose a mathematical framework as well as learning algorithms for finding an effective dynamic treatment regime under model ambiguity.  Incorporating model ambiguity a priori in the analyses not only provides robustness to inevitable misspecifications (e.g., caused by hidden confounders with unknown dynamics and/or  impact on the observed variables), but more broadly can bridge the gap between two philosophical views of causal inference: model-based and model-free.

Our work also tries to close the gap between RL techniques and dynamic causal inference methods.  Specifically, as is common, we view the problem of finding an effective treatment regime as an ``off-policy" RL problem. However, unlike the existing work, we allow the learning to occur across a ``cloud" of potential data generating models. This is specifically useful when data are observational, the behavior policy is unknown, and the existence of time-varying unmeasured confounders (which are themselves affected by previous actions) make the task of learning the causal impact of an evaluation policy challenging.

Unlike the available RL techniques, or the methods related to causal inference in dynamic settings, our work also allows for a two-way personalization: the obtained treatment policies are not only personalized based on the subject's variables (e.g., a patient's covariates), but also based on the ambiguity attitude and preferences of the decision-maker (e.g., the physician). Given the importance of this two-way personalization in a variety of applications (e.g., medical decision-making or public policy), we hope that future research can develop further data-driven methods to learn policies that are personalized in both ways.

We also hope that the future research can test and implement our  prosed learning algorithms in a variety of other applications. In  this study, we investigate the performance of these learning algorithms in three ways.   First, we analytically establish their asymptotic behavior, including (weak) consistency and asymptotic
normality. Second, we examine them in a case study using clinical data related to NODAT patients.  Third, we make use  of simulation experiments (synthetic data), in which we control the true data generating model and compare the performance of our proposed methods with that of an imaginary oracle who knows both the true data generating model and the optimal policy under that model.  All these investigations reveal promising results. However, further research is needed to more broadly investigate the performance of our proposed methods in other applications and domains. With the increasing availability of sensor-based devices that are connected via Internet of Things (IoT) and benefit from data fusion \citep[see, e.g.,][]{SaghafianIoT}, future research can also investigate augmenting our approaches to work with data obtained from multiple connected streams. Finally, future research can examine the interpretability of the policies that are obtained via $\DAV$, $\SAV$, $\DAVBUC$, and $\SAVBUC$, and  propose adjustments (if needed) to ensure that they can be effectively used in practice.

\bibliography{ref}
\bibliographystyle{informs}


%
%




\ECSwitch
\baselineskip=20pt

\ECHead{Online Appendix A: Proofs}
\vspace{2mm}

Available Upon Request

\newpage
\ECHead{Online Appendix B: Supplementary Results and Proofs}
\vspace{2mm}

Available Upon Request

\newpage
\ECHead{Online Appendix C: $\DAVBUC$ and $\SAVBUC$ Algorithms}
\vspace{2mm}

\begin{algorithm}[h]\label{alg:DAVBUC}
\scriptsize
\DontPrintSemicolon
   \For{each observed trajectory  and model $m\in\MM$}
        {
        	Initialize  $\bpi^m_0$ using a random draw from  $F(\bpi)$;  \;
            set t=1;\;
             \While{$t+1\in\MT$}
               {
                 $\bpi^m_{t+1}\leftarrow T(\bpi^m_t, a_t, o_t, m)$;
               }

        }

 \For{any given $\bmu^e\in\Upsilon$ and $m\in\MM$}
 {
  $ \ov\varphi_n^{m,\bmu^e} (\bpsi)\leftarrow \E^{\mathbb P} \Bigg[\sum_{t\in\MT}\bigg[\ov \kappa^m_t\, \frac{\mu^e(A_t|\bPi^m_t)}{\mu^b(A_t|\bPi^m_t)}\Big[G_t+ \beta\,  \ov V_\infty^{m,\bmu^e}  (T(\bPi^m_t, A_t, O_t, m))- \ov V_\infty^{m,\bmu^e} (\bPi^m_t)\Big] \bb (\bPi^m_t)\bigg]\Bigg]$; \;

$ \un\varphi_n^{m,\bmu^e} (\bpsi)\leftarrow \E^{\mathbb P} \Bigg[\sum_{t\in\MT}\bigg[\un \kappa^m_t\, \frac{\mu^e(A_t|\bPi^m_t)}{\mu^b(A_t|\bPi^m_t)}\Big[G_t+ \beta\,  \un V_\infty^{m,\bmu^e}  (T(\bPi^m_t, A_t, O_t, m))- \un V_\infty^{m,\bmu^e} (\bPi^m_t)\Big] \bb (\bPi^m_t)\bigg]\Bigg]$; \;

$\hat{\ov \bpsi}_n^{m,\bmu^e}\leftarrow\arg\!\min_{\bpsi\in\bPsi} \bigg\{\big(\ov\varphi_n^{m,\bmu^e} (\bpsi)\big)'\, \bOmega\, \ov\varphi_n^{m,\bmu^e} (\bpsi)+ \theta_n \mathcal P (\bpsi)\bigg\}$;\;

$\hat{\un \bpsi}_n^{m,\bmu^e}\leftarrow\arg\!\min_{\bpsi\in\bPsi} \bigg\{\big(\un\varphi_n^{m,\bmu^e} (\bpsi)\big)'\, \bOmega\, \un\varphi_n^{m,\bmu^e} (\bpsi)+ \theta_n \mathcal P (\bpsi)\bigg\}$;\;

$\hat{\ov V}^{m,\bmu^e}_{\infty}(\bpi)  \leftarrow \big(\bb (\bpi))'\, \hat{\ov \bpsi}_n^{m,\bmu^e}$;\;

$\hat{\un V}^{m,\bmu^e}_{\infty}(\bpi)  \leftarrow \big(\bb (\bpi))'\, \hat{\un \bpsi}_n^{m,\bmu^e}$;\;

$\hat{\ov\Gamma}^m_{\infty} (\bmu^e) \leftarrow\int  \hat{\ov V}_{\infty}^{m,\bmu^e} (\bpi)\, dF(\bpi)$;\;

$\hat{\un\Gamma}^m_{\infty} (\bmu^e) \leftarrow\int  \hat{\un V}_{\infty}^{m,\bmu^e} (\bpi)\, dF(\bpi)$;\;

}

 \For{any given $\bmu^e\in\Upsilon$}
 {
$\hat{\ov \Gamma}_{\infty} (\bmu^e) \leftarrow \alpha \inf_{m\in\MM} \hat{\ov\Gamma}^m_{\infty} (\bmu^e) + (1-\alpha) \sup_{m\in\MM} \hat{\ov\Gamma}^m_{\infty} (\bmu^e)$;\;

$\hat{\un \Gamma}_{\infty} (\bmu^e) \leftarrow \alpha \inf_{m\in\MM} \hat{\un\Gamma}^m_{\infty} (\bmu^e) + (1-\alpha) \sup_{m\in\MM} \hat{\un\Gamma}^m_{\infty} (\bmu^e)$;\;

$\tilde\alpha \leftarrow \text{TUNE} (\alpha, \epsilon)$;\;

$\hat{\Gamma}_{\infty} (\bmu^e) \leftarrow \tilde\alpha\, \hat{\un \Gamma}_{\infty} (\bmu^e) + (1-\tilde\alpha) \, \hat{\ov \Gamma}_{\infty} (\bmu^e)$;\;

}

$\hat\bmu{^{e*}}  \leftarrow \arg\!\max_{\bmu^e\in\Upsilon} \hat \Gamma_{\infty} (\bmu^e)$;\;

$\hat \Gamma_{\infty} (\hat\bmu^{e*} )\leftarrow \max_{\bmu^e\in\Upsilon} \hat \Gamma_{\infty} (\bmu^e)$;

\caption{$\DAVBUC$}
\end{algorithm}

\begin{algorithm}[h]\label{alg:SAVBUC}
\scriptsize
\DontPrintSemicolon
   \For{each observed trajectory  and model $m\in\MM$}
        {
        	Initialize  $\bpi^m_0$ using a random draw from  $F(\bpi)$;  \;
            set t=1;\;
             \While{$t+1\in\MT$}
               {
                 $\bpi^m_{t+1}\leftarrow T(\bpi^m_t, a_t, o_t, m)$;
               }

        }

 \For{any given $\bmu^e\in\Upsilon$ and $m\in\MM$}
 {
  $ \ov\varphi_n^{m,\bmu^e} (\bpsi)\leftarrow \E^{\mathbb P} \Bigg[\sum_{t\in\MT}\bigg[\ov \kappa^m_t\, \frac{\mu^e(A_t|\bPi^m_t)}{\mu^b(A_t|\bPi^m_t)}\Big[G_t+ \beta\,  \ov V_\infty^{m,\bmu^e}  (T(\bPi^m_t, A_t, O_t, m))- \ov V_\infty^{m,\bmu^e} (\bPi^m_t)\Big] \bb (\bPi^m_t)\bigg]\Bigg]$; \;

$ \un\varphi_n^{m,\bmu^e} (\bpsi)\leftarrow \E^{\mathbb P} \Bigg[\sum_{t\in\MT}\bigg[\un \kappa^m_t\, \frac{\mu^e(A_t|\bPi^m_t)}{\mu^b(A_t|\bPi^m_t)}\Big[G_t+ \beta\,  \un V_\infty^{m,\bmu^e}  (T(\bPi^m_t, A_t, O_t, m))- \un V_\infty^{m,\bmu^e} (\bPi^m_t)\Big] \bb (\bPi^m_t)\bigg]\Bigg]$; \;

$\hat{\ov \bpsi}_n^{m,\bmu^e}\leftarrow\arg\!\min_{\bpsi\in\bPsi} \bigg\{\big(\ov\varphi_n^{m,\bmu^e} (\bpsi)\big)'\, \bOmega\, \ov\varphi_n^{m,\bmu^e} (\bpsi)+ \theta_n \mathcal P (\bpsi)\bigg\}$;\;

$\hat{\un \bpsi}_n^{m,\bmu^e}\leftarrow\arg\!\min_{\bpsi\in\bPsi} \bigg\{\big(\un\varphi_n^{m,\bmu^e} (\bpsi)\big)'\, \bOmega\, \un\varphi_n^{m,\bmu^e} (\bpsi)+ \theta_n \mathcal P (\bpsi)\bigg\}$;\;
}

 \For{any given $\bmu^e\in\Upsilon$}
 {

$\un m_1 \leftarrow \arg\!\inf_{m\in\MM} ||\hat{\ov \bpsi}_n^{m,\bmu^e}||$; \;

$\ov m_1 \leftarrow \arg\!\sup_{m\in\MM} ||\hat{\ov \bpsi}_n^{m,\bmu^e}||$; \;

$\un m_2 \leftarrow \arg\!\inf_{m\in\MM} ||\hat{\un \bpsi}_n^{m,\bmu^e}||$; \;

$\ov m_2 \leftarrow \arg\!\sup_{m\in\MM} ||\hat{\un \bpsi}_n^{m,\bmu^e}||$; \;

$\hat{\ov\bpsi}^{\bmu^e}_n \leftarrow \alpha \, \hat{\ov\bpsi}_n^{\un m_1,\bmu^e}+ (1-\alpha)\, \hat{\ov\bpsi}_n^{\ov m_1,\bmu^e}$; \;

$\hat{\un\bpsi}^{\bmu^e}_n \leftarrow \alpha \, \hat{\un\bpsi}_n^{\un m_2,\bmu^e}+ (1-\alpha)\, \hat{\un\bpsi}_n^{\ov m_2,\bmu^e}$; \;

$\hat{\ov V}^{\bmu^e}_{\infty}(\bpi)  \leftarrow \big(\bb (\bpi))'\, \hat{\ov\bpsi}_n^{\bmu^e}$;\;

$\hat{\un V}^{\bmu^e}_{\infty}(\bpi)  \leftarrow \big(\bb (\bpi))'\, \hat{\un\bpsi}_n^{\bmu^e}$;\;

$\hat{ \ov \Gamma}_{\infty} (\bmu^e) \leftarrow \int  \hat{\ov V}_{\infty}^{\bmu^e} (\bpi)\, dF(\bpi)$;\;

$\hat{ \un \Gamma}_{\infty} (\bmu^e) \leftarrow \int  \hat{\un V}_{\infty}^{\bmu^e} (\bpi)\, dF(\bpi)$;\;

$\tilde\alpha \leftarrow \text{TUNE} (\alpha, \epsilon)$;\;

$\hat{\Gamma}_{\infty} (\bmu^e) \leftarrow \tilde\alpha\, \hat{\un \Gamma}_{\infty} (\bmu^e) + (1-\tilde\alpha) \, \hat{\ov \Gamma}_{\infty} (\bmu^e)$;\;

}
$\hat\bmu{^{e*}}  \leftarrow \arg\!\max_{\bmu^e\in\Upsilon} \hat \Gamma_{\infty} (\bmu^e)$;\;

$\hat \Gamma_{\infty} (\hat\bmu^{e*} )\leftarrow \max_{\bmu^e\in\Upsilon} \hat \Gamma_{\infty} (\bmu^e)$;

\caption{$\SAVBUC$}
\end{algorithm}


\newpage
\ECHead{Online Appendix D: Additional Figures}
\vspace{2mm}

 \end{document}